\documentclass[10pt,journal]{IEEEtran}

\usepackage[hidelinks]{hyperref}

\usepackage{graphicx}

\usepackage{mathtools}

\usepackage{amsfonts}

\usepackage{amssymb}

\usepackage{algorithm}

\usepackage[noend]{algpseudocode}

\usepackage{nicefrac}

\usepackage[keeplastbox]{flushend}

\usepackage{caption}

\newcommand{\fig}[1]{Fig.~\ref{#1}}

\newcommand{\eq}[1]{(\ref{#1})}

\newcommand{\tbl}[1]{(Table~\ref{#1})}

\newcommand{\alg}[1]{Algorithm~\ref{#1}}

\algrenewcommand{\algorithmiccomment}[1]{\# #1}

\newlength{\argwd}
\newlength{\arght}
\newcommand{\overharp}[3]{%
	\settowidth{\argwd}{#2}%
	\settoheight{\arght}{#2}%
	\addtolength{\argwd}{.1\argwd}%
	\raisebox{\arght}{%
		\makebox[.04\argwd][l]{%
			\resizebox{\argwd}{#3\arght}{$#1$}%
		}%
	}%
	#2%
}
\newcommand{\overrightharp}[2]{\overharp{\rightharpoonup}{#1}{#2}}
\newcommand{\vect}[2][.5]{\text{\overrightharp{\ensuremath{\boldsymbol{#2}}}{#1}}}
\newcommand{\vectmd}[2][.5]{\text{\overrightharp{\ensuremath{#2}}{#1}}}

\newsavebox{\mybox}\newsavebox{\mysim}
\newcommand{\distas}[1]{%
  \savebox{\mybox}{\hbox{$\scriptstyle#1$}}%
  \savebox{\mysim}{\hbox{$\sim$}}%
  \mathbin{\overset{#1}{\resizebox{\wd\mybox}{\ht\mysim}{$\sim$}}}%
}

\begin{document}
	\title{A Mathematical Formalization of Hierarchical Temporal Memory's Spatial Pooler}
	
	\author{
        James~Mnatzaganian,~\IEEEmembership{Student~Member,~IEEE,}
		Ernest~Fokou{\'e},~\IEEEmembership{Member,~IEEE}
        and~Dhireesha~Kudithipudi,~\IEEEmembership{Senior~Member,~IEEE}%
		
		\thanks{This work began in August 2014.}%
		\thanks{J. Mnatzaganian and D. Kudithipudi are with the NanoComputing Research Laboratory, Rochester Institute of Technology, Rochester, NY, 14623.}%
		\thanks{E. Fokou{\'e} is with the Data Science Research Group, Rochester Institute of Technology, Rochester, NY, 14623.
		}
	}
	
	\markboth{ }%
	{Mnatzaganian \MakeLowercase{\textit{et al.}}: Mathematical Formalization of Hierarchical Temporal Memory Cortical Learning Algorithm's Spatial Pooler}
	
	\maketitle
	
	\begin{abstract}
		Hierarchical temporal memory~(HTM) is an emerging machine learning algorithm, with the potential to provide a means to perform predictions on spatiotemporal data. The algorithm, inspired by the neocortex, currently does not have a comprehensive mathematical framework. This work brings together all aspects of the spatial pooler~(SP), a critical learning component in HTM, under a single unifying framework. The primary learning mechanism is explored, where a maximum likelihood estimator for determining the degree of permanence update is proposed. The boosting mechanisms are studied and found to be a secondary learning mechanism. The SP is demonstrated in both spatial and categorical multi-class classification, where the SP is found to perform exceptionally well on non-spatial data. Observations are made relating HTM to well-known algorithms such as competitive learning and attribute bagging. Methods are provided for using the SP for classification as well as dimensionality reduction. Empirical evidence verifies that given the proper parameterizations, the SP may be used for feature learning. 
	\end{abstract}
	
		
	\section{Introduction}		
		\IEEEPARstart{H}{ierarchical} temporal memory~(HTM) is a machine learning algorithm that was inspired by the neocortex and designed to learn sequences and make predictions. In its idealized form, it should be able to produce generalized representations for similar inputs. Given time-series data, HTM should be able to use its learned representations to perform a type of time-dependent regression. Such a system would prove to be incredibly useful in many applications utilizing spatiotemporal data. One instance for using HTM with time-series data was recently demonstrated by Cui~et~al.~\cite{numenta_tp_prediction}, where HTM was used to predict taxi passenger counts. The use of HTM in other applications remains unexplored, largely due to the evolving nature of HTM's algorithmic definition. Additionally, the lack of a formalized mathematical model hampers its prominence in the machine learning community. This work aims to bridge the gap between a neuroscience inspired algorithm and a math-based algorithm by constructing a purely mathematical framework around HTM's original algorithmic definition.
		
\begin{figure}[!t]
			\centering
			\includegraphics[width=\linewidth]{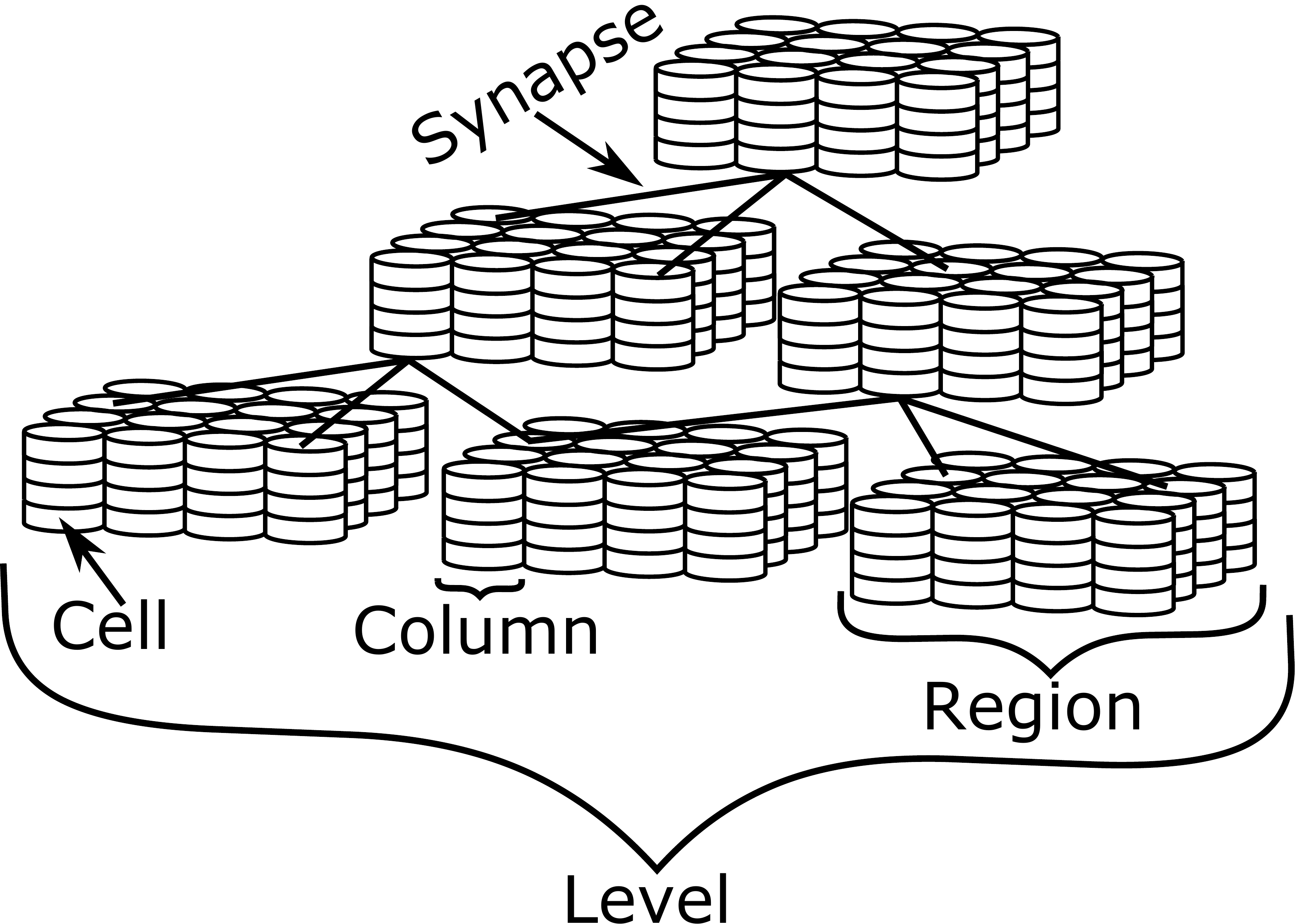}
			\caption{Depiction of HTM, showing the various levels of detail.}
			\label{fig:htm}
		\end{figure}

		HTM models, at a high-level, some of the structures and functionality of the neocortex. Its structure follows that of cortical minicolumns, where an HTM region is comprised of many columns, each consisting of multiple cells. One or more regions form a level. Levels are stacked hierarchically in a tree-like structure to form the full network depicted in \fig{fig:htm}. Within HTM, connections are made via synapses, where both proximal and distal synapses are utilized to form feedforward and neighboring connections, respectively.
		
		The current version of HTM is the predecessor to HTM cortical learning algorithm~(CLA)~\cite{cla_whitepaper}. In the current version of HTM the two primary algorithms are the spatial pooler~(SP) and the temporal memory~(TM). The SP is responsible for taking an input, in the format of a sparse distributed representation (SDR), and producing a new SDR. In this manner, the SP can be viewed as a mapping function from the input domain to a new feature domain. In the feature domain a single SDR should be used to represent similar SDRs from the input domain. The algorithm is a type of unsupervised competitive learning algorithm that uses a form of vector quantization~(VQ) resembling self-organizing maps~(SOMs). The TM is responsible for learning sequences and making predictions. This algorithm follows Hebb's rule~\cite{hebb}, where connections are formed between cells that were previously active. Through the formation of those connections a sequence may be learned. The TM can then use its learned knowledge of the sequences to form predictions.
		
		HTM originated as an abstraction of the neocortex; as such, it does not have an explicit mathematical formulation. Without a mathematical framework, it is difficult to understand the key characteristics of the algorithm and how it can be improved. In general, very little work exists regarding the mathematics behind HTM. Hawkins~et~al.~\cite{numneta_tp_math} recently provided a starting mathematical formulation for the TM, but no mentions to the SP were made. Lattner~\cite{lattner} provided an initial insight about the SP, by relating it to VQ. He additionally provided some equations governing computing overlap and performing learning; however, those equations were not generalized to account for local inhibition. Byrne~\cite{byrne} began the use of matrix notation and provided a basis for those equations; however, certain components of the algorithm, such as boosting, were not included. Leake~et~al.~\cite{leake} provided some insights regarding the initialization of the SP. He also provided further insights into how the initialization may affect the initial calculations within the network; however, his focus was largely on the network initialization. The goal of this work is to provide a complete mathematical framework for HTM's SP and to demonstrate how it may be used in various machine learning tasks.
		
		The major, novel contributions provided by this work are as follows:
		\begin{itemize}
			\item Creation of a complete mathematical framework for the SP, including boosting and local inhibition.
			\item Using the SP to perform feature learning.
			\item Using the SP as a pre-processor for non-spatial data.
			\item Creation of a possible mathematical explanation for the permanence update amount.
			\item Insights into the permanence selection.
		\end{itemize}
	
	\section{Spatial Pooler Algorithm}
		The SP consists of three phases, namely overlap, inhibition, and learning. In this section the three phases will be presented based off their original, algorithmic definition. This algorithm follows an iterative, online approach, where the learning updates occur after the presentation of each input. Before the execution of the algorithm, some initializations must take place.
		
		Within an SP there exist many columns. Each column has a unique set of proximal synapses connected via a proximal dendrite segment. Each proximal synapse tentatively connects to a single column from the input space, where the column's activity level is used as an input, i.e. an active column is a `1' and an inactive column is a `0'.
		
		To determine whether a synapse is connected or not, the synapse's permanence value is checked. If the permanence value is at least equal to the connected threshold the synapse is connected; otherwise, it is unconnected. The permanence values are scalars in the closed interval [0, 1].
		
		\begin{algorithm}[t]
			\caption{SP phase 1: Overlap}
			\label{alg:phase1}
			\begin{algorithmic}[1]
				\ForAll {$col \in sp.columns$}
					\State $col.overlap \leftarrow 0$
					\ForAll {$syn \in col.\operatorname{connected\_synapses}()$}
						\State $col.overlap \leftarrow col.overlap + syn.\operatorname{active}()$
					\EndFor
					
					\Statex
					\If {$col.overlap < pseg\_th$}
						\State $col.overlap \leftarrow 0$
					\Else
						\State $col.overlap \leftarrow col.overlap * col.boost$
					\EndIf
				\EndFor
			\end{algorithmic}
		\end{algorithm}
		
		Prior to the first execution of the algorithm, the potential connections of proximal synapses to the input space and the initial permanence values must be determined. Following Numenta's whitepaper~\cite{cla_whitepaper}, each synapse is randomly connected to a unique input bit, i.e. the number of synapses per column and the number of input bits are binomial coefficients. The permanences of the synapses are then randomly initialized to a value close to the connected permanence threshold. A second constraint requires that the permanence value be a function of the distance between the SP column's position and the input column's position, such that the closer the input is to the column the larger the value should be. The three phases of the SP are explained in the following subsections.
		
		\subsection{Phase 1: Overlap}
			The first phase of the SP is used to compute the overlap between each column and its respective input, as shown in \alg{alg:phase1}. In \alg{alg:phase1}, the SP is represented by the object $sp$. The method $col.\operatorname{connected\_synapses}()$ returns an instance to each synapse on $col$'s proximal segment that is connected, i.e. synapses having permanence values greater than the permanence connected threshold, $psyn\_th$. The method $syn.\operatorname{active}()$ returns `1' if $syn$'s input is active and `0' otherwise. $pseg\_th$\footnote{This parameter was originally referred to as the minimum overlap; however, it is renamed in this work to allow consistency between the SP and the TM.} is a parameter that determines the activation threshold of a proximal segment, such that there must be at least $pseg\_th$ active connected proximal synapses on a given proximal segment for it to become active. The parameter $col.boost$ is the boost value for $col$, which is initialized to `1' and updated according to \alg{alg:boost}.
		
		\subsection{Phase 2: Inhibition}
			The second phase of the SP is used to compute the set of active columns after they have been inhibited, as shown in \alg{alg:phase2}. In \alg{alg:phase2}, $\operatorname{kmax\_overlap}(C, k)$ is a function that returns the $k$-th largest overlap of the columns in $C$. The method $sp.\operatorname{neighbors}(col)$ returns the columns that are within $col$'s neighborhood, including $col$, where the size of the neighborhood is determined by the inhibition radius. The parameter $k$ is the desired column activity level. In line 2 in \alg{alg:phase2}, the $k$-th largest overlap value out of $col$'s neighborhood is being computed. A column is then said to be active if its overlap value is greater than zero and the computed minimum overlap, $mo$.
			
			\begin{algorithm}[t]
				\caption{SP phase 2: Inhibition}
				\label{alg:phase2}
				\begin{algorithmic}[1]
					\ForAll {$col \in sp.columns$}
						\State $mo \leftarrow \operatorname{kmax\_overlap}(sp.\operatorname{neighbors}(col), k)$
						
						\Statex
						\If {$col.overlap > 0$ \textbf{and} $col.overlap \geq mo$}
							\State $col.active \leftarrow 1$
						\Else
							\State $col.active \leftarrow 0$
						\EndIf
					\EndFor
				\end{algorithmic}
			\end{algorithm}
		
		\subsection{Phase 3: Learning}
			The third phase of the SP is used to conduct the learning operations, as shown in \alg{alg:phase3}. This code contains three parts -- permanence adaptation, boosting operations, and the inhibition radius update. In \alg{alg:phase3}, $syn.p$ refers to the permanence value of $syn$. The functions $\operatorname{min}$ and $\operatorname{max}$ return the minimum and maximum values of their arguments, respectively, and are used to keep the permanence values bounded in the closed interval [0, 1]. The constants $syn.psyn\_inc$ and $syn.psyn\_dec$ are the proximal synapse permanence increment and decrement amounts, respectively.
			
			The function $\operatorname{max\_adc}(C)$ returns the maximum active duty cycle of the columns in $C$, where the active duty cycle is a moving average denoting the frequency of column activation. Similarly, the overlap duty cycle is a moving average denoting the frequency of the column's overlap value being at least equal to the proximal segment activation threshold. The functions $col.\operatorname{update\_active\_duty\_cycle}()$ and $col.\operatorname{update\_overlap\_duty\_cycle}()$ are used to update the active and overlap duty cycles, respectively, by computing the new moving averages. The parameters $col.odc$, $col.adc$, and $col.mdc$ refer to $col$'s overlap duty cycle, active duty cycle, and minimum duty cycle, respectively. Those duty cycles are used to ensure that columns have a certain degree of activation. 
			
			The method $col.\operatorname{update\_boost}()$ is used to update the boost for column, $col$, as shown in \alg{alg:boost}, where $maxb$ refers to the maximum boost value. It is important to note that the whitepaper did not explicitly define how the boost should be computed. This boost function was obtained from the source code of Numenta's implementation of HTM, Numenta platform for intelligent computing (NuPIC)~\cite{nupic}.
			
			The method $sp.\operatorname{update\_inhibition\_radius}()$ is used to update the inhibition radius. The inhibition radius is set to the average receptive field size, which is average distance between all connected synapses and their respective columns in the input and the SP.
			
			\begin{algorithm}[t]
				\caption{SP phase 3: Learning}
				\label{alg:phase3}
				\begin{algorithmic}[1]
					\Statex \Comment Adapt permanences
					\ForAll {$col \in sp.columns$}
						\If {$col.active$}
							\ForAll {$syn \in col.synapses$}
								\If {$syn.\operatorname{active}()$}
									\State {$syn.p \leftarrow \operatorname{min}(1, syn.p + syn.psyn\_inc)$}
								\Else
									\State {$syn.p \leftarrow \operatorname{max}(0, syn.p - syn.psyn\_dec)$}
								\EndIf
							\EndFor
						\EndIf
					\EndFor
					
					\Statex
					\Statex \Comment Perform boosting operations
					\ForAll {$col \in sp.columns$}
						\State {$col.mdc \leftarrow 0.01 * \operatorname{max\_adc}(sp.\operatorname{neighbors}(col))$}
						\State {$col.\operatorname{update\_active\_duty\_cycle}()$}
						\State {$col.\operatorname{update\_boost}()$}
						
						\Statex
						\State {$col.\operatorname{update\_overlap\_duty\_cycle}()$}
						\If {$col.odc < col.mdc$}
							\ForAll {$syn \in col.synapses$}
								\State {$syn.p \leftarrow \operatorname{min}(1, syn.p + 0.1 * psyn\_th)$}
							\EndFor
						\EndIf
					\EndFor
					
					\Statex
					\State {$sp.\operatorname{update\_inhibition\_radius}()$}
				\end{algorithmic}
			\end{algorithm}
	
	\section{Mathematical Formalization}
		The aforementioned operation of the SP lends itself to a vectorized notation. By redefining the operations to work with vectors it is possible not only to create a mathematical representation, but also to greatly improve upon the efficiency of the operations. The notation described in this section will be used as the notation for the remainder of the document.
		
		All vectors will be lowercase, bold-faced letters with an arrow hat. Vectors are assumed to be row vectors, such that the transpose of the vector will produce a column vector.	All matrices will be uppercase, bold-faced letters.	Subscripts on vectors and matrices are used to denote where elements are being indexed, following a row-column convention, such that $\boldsymbol{X}_{i,j} \in \boldsymbol{X}$ refers to $\boldsymbol{X}$ at row index\footnote{All indices start at 0.} $i$ and column index $j$. Element-wise operations between a vector and a matrix are performed column-wise, such that $\vect{x}^T \odot \boldsymbol{Y} = \vect{x}_i\boldsymbol{Y}_{i,j}\ \forall i\ \forall j$.
		
		Let $\operatorname{I}(k)$ be defined as the indicator function, such that the function will return 1 if event $k$ is true and 0 otherwise. If the input to this function is a vector of events or a matrix of events, each event will be evaluated independently, with the function returning a vector or matrix of the same size as its input. Any variable with a superscript in parentheses is used to denote the type of that variable. For example, $\vect{x}^{(y)}$ is used to state that the variable $\vect{x}$ is of type $y$.
		
		\begin{algorithm}[t]
			\caption{Boost Update: $col.\operatorname{update\_boost}()$}
			\label{alg:boost}
			\begin{algorithmic}[1]
				\If {$col.mdc == 0$}
					\State $col.boost \leftarrow maxb$
				\ElsIf {$col.adc > col.mdc$}
					\State $col.boost \leftarrow 1$
				\Else
					\State $col.boost = col.adc * ((1 - maxb) / col.mdc) + maxb$
				\EndIf
			\end{algorithmic}
		\end{algorithm}
		
		\begin{table}[!t]
			\caption{User-defined parameters for the SP}
			\label{tbl:parameters}
			\centering
			\begin{tabular}{l|l}
				\hline
				\multicolumn{1}{c|}{Parameter} & \multicolumn{1}{c}{Description} \\
				\hline
				$n$ & Number of patterns (samples) \\
				$p$ & Number of inputs (features) in a pattern \\
				$m$ & Number of columns \\
				$q$ & Number of proximal synapses per column \\
				$\phi_+$ & Permanence increment amount \\
				$\phi_-$ & Permanence decrement amount \\
				$\phi_{\delta}$ & Window of permanence initialization \\
				$\rho_d$ & Proximal dendrite segment activation threshold \\
				$\rho_s$ & Proximal synapse activation threshold \\
				$\rho_c$ & Desired column activity level \\
				$\kappa_a$ & Minimum activity level scaling factor \\
				$\kappa_b$ & Permanence boosting scaling factor \\
				$\beta_0$ & Maximum boost \\
				$\tau$ & Duty cycle period \\
				\hline
			\end{tabular}
		\end{table}
		
		All of the user-defined parameters are defined in \tbl{tbl:parameters}\footnote{The parameters $\kappa_a$ and $\kappa_b$ have default values of 0.01 and 0.1, respectively.}. These are parameters that must be defined before the initialization of the algorithm. All of those parameters are constants, except for parameter $\rho_c$, which is an overloaded parameter. It can either be used as a constant, such that for a column to be active it must be greater than the $\rho_c$-th column's overlap. It may also be defined to be a density, such that for a column to be active it must be greater than the $\lfloor \rho_c * \operatorname{num\_neighbors}(i) \rfloor$-th column's overlap, where $\operatorname{num\_neighbors}(i)$ is a function that returns the number of neighbors that column $i$ has. If $\rho_c$ is an integer it is assumed to be a constant, and if it is a scalar in the interval (0, 1] it is assumed to be used as a density.
		
		Let the terms $s$, $r$, $i$, $j$, and $k$ be defined as integer indices. They are henceforth bounded as follows: $s \in [0, n)$, $r \in [0, p)$, $i \in [0, m)$, $j \in [0, m)$, and $k \in [0, q)$.
		
		\subsection{Initialization}
			Competitive learning networks typically have each node fully connected to each input. The SP; however, follows a different line of logic, posing a new problem concerning the visibility of an input. As previously explained, the inputs connecting to a particular column are determined randomly. Let $\vect{c} \in \mathbb{Z}^{1 \times m}, \vect{c} \in [0, m)$ be defined as the set of all columns indices, such that $\vect{c}_i$ is the column's index at $i$. Let $\boldsymbol{U} \in \{0, 1\}^{n \times p}$ be defined as the set of inputs for all patterns, such that $\boldsymbol{U}_{s, r}$ is the input for pattern $s$ at index $r$. Let $\boldsymbol{\Lambda} \in \{r\}^{m \times q}$ be the source column indices for each proximal synapse on each column, such that $\boldsymbol{\Lambda}_{i,k}$ is the source column's index of $\vect{c}_i$'s proximal synapse at index $k$. In other words, each $\boldsymbol{\Lambda}_{i,k}$ refers to a specific index in $\boldsymbol{U}_s$.
			
			Let $\vect[0.4]{ic}_r\equiv\exists! r\in\boldsymbol{\Lambda}_i\ \forall r$, the event of input $r$ connecting to column $i$, where $\exists!$ is defined to be the uniqueness quantification. Given $q$ and $p$, the probability of a single input, $\boldsymbol{U}_{s, r}$, connecting to a column is calculated by using \eq{eq:p_connect}. In \eq{eq:p_connect}, the probability of an input not connecting is first determined. That probability is independent for each input; thus, the total probability of a connection not being formed is simply the product of those probabilities. The probability of a connection forming is therefore the complement of the probability of a connection not forming.
			
			\begin{equation}
				\begin{split}
					\mathbb{P}(\vect{ic}_r) &= 1 - \prod_{k=0}^{q}\left(1 - \frac{1}{p - k}\right)\\
					&=\frac{q+1}{p}
				\end{split}
				\label{eq:p_connect}
			\end{equation}
			
			It is also desired to know the average number of columns an input will connect with. To calculate this, let $\vect[0.4]{\lambda}\equiv \sum_{i=0}^{m-1}\sum_{k=0}^{q-1}\operatorname{I}(r=\Lambda_{i,k})\ \forall r$, the random vector governing the count of connections between each input and all columns. Recognizing that the probability of a connection forming in $m$ follows a binomial distribution, the expected number of columns that an input will connect to is simply \eq{eq:e_columns}.
			
			\begin{equation}
				\mathbb{E}\left[\vect{\lambda}_r\right] = m\mathbb{P}(\vect{ic}_r)
				\label{eq:e_columns}
			\end{equation}
			
			Using \eq{eq:p_connect} it is possible to calculate the probability of an input never connecting, as shown in \eq{eq:p_never_connect}. Since the probabilities are independent, it simply reduces to the product of the probability of an input not connecting to a column, taken over all columns. Let $\lambda'\equiv \sum_{r=0}^{p-1}\operatorname{I}(\vect[0.4]{\lambda}_r=0)$, the random variable governing the number of unconnected inputs. From \eq{eq:p_never_connect}, the expected number of unobserved inputs may then be trivially obtained as \eq{eq:e_unobserved}. Using \eq{eq:p_never_connect} and \eq{eq:e_columns}, it is possible to obtain a lower bound for $m$ and $q$, by choosing those parameters such that a certain amount of input visibility is obtained. To guarantee observance of all inputs, \eq{eq:p_never_connect} must be zero. Once that is satisfied, the desired number of times an input is observed may be determined by using \eq{eq:e_columns}.
			
			\begin{equation}
				\mathbb{P}\left(\vect{\lambda}_r=0\right) = (1-\mathbb{P}(\vect{ic}_r))^{m}
				\label{eq:p_never_connect}
			\end{equation}
			
			\begin{equation}
				\mathbb{E}[\lambda'] = p\mathbb{P}\left(\vect{\lambda}_r=0\right)
				\label{eq:e_unobserved}
			\end{equation}
			
			Once each column has its set of inputs, the permanences must be initialized. As previously stated, permanences were defined to be initialized with a random value close to $\rho_s$, but biased based off the distance between the synapse's source (input column) and destination (SP column). To obtain further clarification, NuPIC's source code~\cite{nupic} was consulted. It was found that the permanences were randomly initialized, with approximately half of the permanences creating connected proximal synapses and the remaining permanences creating potential (unconnected) proximal synapses. Additionally, to ensure that each column has a fair chance of being selected during inhibition, there are at least $\rho_d$ connected proximal synapses on each column.
			
			Let $\boldsymbol{\Phi} \in \mathbb{R}^{m \times q}$ be defined as the set of permanences for each column, such that $\boldsymbol{\Phi}_i$ is the set of permanences for the proximal synapses for $\vect{c}_i$. Each $\boldsymbol{\Phi}_{i,k}$ is randomly initialized as shown in \eq{eq:perm_init}, where $\operatorname{Unif}$ represents the uniform distribution. Using \eq{eq:perm_init}, the expected permanence value would be equal to $\rho_s$; thus, $\nicefrac{q}{2}$ proximal synapses would be initialized as connected for each column. To ensure that each column has a fair chance of being selected, $\rho_d$ should be less than $\nicefrac{q}{2}$.
			
			\begin{equation}
				\boldsymbol{\Phi}_{i,k} \sim \operatorname{Unif}(\rho_s - \phi_{\delta}, \rho_s + \phi_{\delta})
				\label{eq:perm_init}
			\end{equation}
			
			It is possible to predict, before training, the initial response of the SP with a given input. This insight allows parameters to be crafted in a manner that ensures a desired amount of column activity. Let $\boldsymbol{X} \in \{0, 1\}^{m \times q}$ be defined as the set of inputs for each column, such that $\boldsymbol{X}_i$ is the set of inputs for $\vect{c}_i$. Let $\vect{ai}_i\equiv\sum_{k=0}^{q-1}\boldsymbol{X}_{i,k}$, the random variable governing the number of active inputs on column $i$. Let $\mathbb{P}(\boldsymbol{X}_{i,k})$ be defined as the probability of the input connected via proximal synapse $k$ to column $i$ being active. $\mathbb{P}(\boldsymbol{X}_i)$ is therefore defined to be the probability of an input connected to column $i$ being active. Similarly, $\mathbb{P}(\boldsymbol{X})$ is defined to be the probability of an input on any column being active. The expected number of active proximal synapses on column $i$ is then given by \eq{eq:e_active}. Let $a\equiv\frac{1}{m}\sum_{i=0}^{m-1}\sum_{k=0}^{q-1}\boldsymbol{X}_{i,k}$, the random variable governing the average number of active inputs on a column. Equation \eq{eq:e_active} is then generalized to \eq{eq:e_active_generalized}, the expected number of active proximal synapses for each column.
						
			\begin{equation}
				\mathbb{E}[\vect{ai}_i] = q\mathbb{P}(\boldsymbol{X}_i)
				\label{eq:e_active}
			\end{equation}
									
			\begin{equation}
				\mathbb{E}[a] = q\mathbb{P}(\boldsymbol{X})
				\label{eq:e_active_generalized}
			\end{equation}
			
			Let $\boldsymbol{AC}_{i,k}\equiv\boldsymbol{X}_{i,k}\cap \operatorname{I}\left(\boldsymbol{\Phi}_{i,k}\ge \rho_s\right)$, the event that proximal synapse $k$ is active and connected on column $i$. Let $\vect{ac}_i\equiv\sum_{k=0}^{q-1}\boldsymbol{AC}_{i,k}$, the random variable governing the number of active and connected proximal synapses for column $i$. Let $\mathbb{P}(\boldsymbol{AC}_{i,k})\equiv\mathbb{P}(\boldsymbol{X}_{i,k})\rho_s$, the probability that a proximal synapse is active and connected\footnote{$\rho_s$ was used as a probability. Because $\rho_s\in\mathbb{R}, \rho_s\in(0,1)$, permanences are uniformly initialized with a mean of $\rho_s$, and for a proximal synapse to be connected it must have a permanence value at least equal to $\rho_s$, $\rho_s$ may be used to represent the probability that an initialized proximal synapse is connected.}. Following \eq{eq:e_active}, the expected number of active connected proximal synapses on column $i$ is given by \eq{eq:e_active_connected}.
			
			\begin{equation}
				\mathbb{E}[\vect{ac}_i] = q\mathbb{P}(\boldsymbol{AC}_{i,k})
				\label{eq:e_active_connected}
			\end{equation}
			
			Let $\operatorname{Bin}(k; n, p)$ be defined as the probability mass function (PMF) of a binomial distribution, where $k$ is the number of successes, $n$ is the number of trials, and $p$ is the success probability in each trial. Let $at\equiv\sum_{i=0}^{m-1}\operatorname{I}\left(\left(\sum_{k=0}^{q-1}\boldsymbol{X}_{i,k}\right) \ge \rho_d\right)$, the random variable governing the number of columns having at least $\rho_d$ active proximal synapses. Let $act\equiv\sum_{i=0}^{m-1}\operatorname{I}\left(\left(\sum_{k=0}^{q-1}\boldsymbol{AC}_{i,k}\right) \ge \rho_d\right)$, the random variable governing the number of columns having at least $\rho_d$ active connected proximal synapses. Let $\pi_x$ and $\pi_{ac}$ be defined as random variables that are equal to the overall mean of $\mathbb{P}(\boldsymbol{X})$ and $\mathbb{P}(\boldsymbol{AC})$, respectively. The expected number of columns with at least $\rho_d$ active proximal synapses and the expected number of columns with at least $\rho_d$ active connected proximal synapses are then given by \eq{eq:e_active_threshold} and \eq{eq:e_active_connected_threshold}, respectively.
			
			In \eq{eq:e_active_threshold}, the summation computes the probability of having less than $\rho_d$ active connected proximal synapses, where the individual probabilities within the summation follow the PMF of a binomial distribution. To obtain the desired probability, the complement of that probability is taken. It is then clear that the mean is nothing more than that probability multiplied by $m$. For \eq{eq:e_active_connected_threshold} the logic is similar, with the key difference being that the probability of a success is a function of both $\boldsymbol{X}$ and $\rho_s$, as it was in \eq{eq:e_active_connected}.
			
			\begin{figure}[!t]
				\centering
				\includegraphics[width=\linewidth]{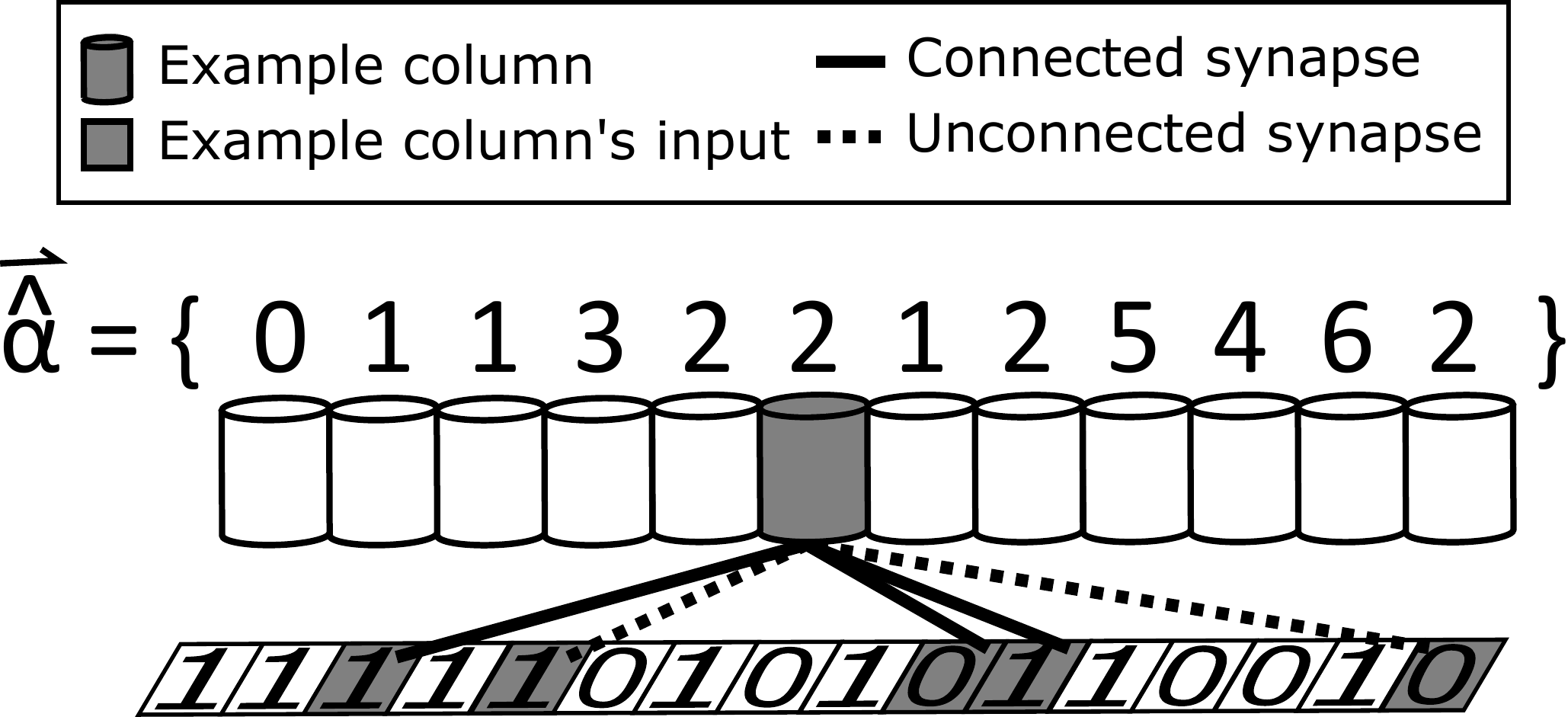}
				\caption{SP phase 1 example where $m = 12$, $q = 5$, and $\rho_d = 2$. It was assumed that the boost for all columns is at the initial value of `1'. For simplicity, only the connections for the example column, highlighted in gray, are shown.}
				\label{fig:phase1}
			\end{figure}
			
			\begin{equation}
				\mathbb{E}[at]=m\left[1-\sum_{t=0}^{\rho_d-1}\operatorname{Bin}(t; q, \pi_x)\right]
				\label{eq:e_active_threshold}
			\end{equation}
			
			\begin{equation}
				\mathbb{E}[act]=m\left[1-\sum_{t=0}^{\rho_d-1}\operatorname{Bin}(t; q, \pi_{ac})\right]
				\label{eq:e_active_connected_threshold}
			\end{equation}
		
		\subsection{Phase 1: Overlap}
			Let $\vect[0.3]{b} \in \mathbb{R}^{1 \times m}$ be defined as the set of boost values for all columns, such that $\vect[0.3]{b}_i$ is the boost for $\vect{c}_i$. Let $\boldsymbol{Y}\equiv\operatorname{I}(\boldsymbol{\Phi}_i \geq \rho_s)\ \forall i$, the bit-mask for the proximal synapse's activations. $\boldsymbol{Y}_i$ is therefore a row-vector bit-mask, with each `1' representing a connected synapse and each `0' representing an unconnected synapse. In this manner, the connectivity (or lack thereof) for each synapse on each column is obtained. The overlap for all columns, $\vect{\alpha} \in \{0, 1\}^{1 \times m}$, is then obtained by using \eq{eq:overlap}, which is a function of $\vect[0.3]{\hat{\alpha}} \in \mathbb{Z}^{1 \times m}$. $\vect[0.3]{\hat{\alpha}}$ is the sum of the active connected proximal synapses for all columns, and is defined in \eq{eq:preoverlap}.
			
			Comparing these equations with \alg{alg:phase1}, it is clear that $\vect[0.3]{\hat{\alpha}}$ will have the same value as $col.overlap$ before line five, and that the final value of $col.overlap$ will be equal to $\vect{\alpha}$. To help provide further understanding, a simple example demonstrating the functionality of this phase is shown in \fig{fig:phase1}.
			
			\begin{figure}[!t]
				\centering
				\includegraphics[width=\linewidth]{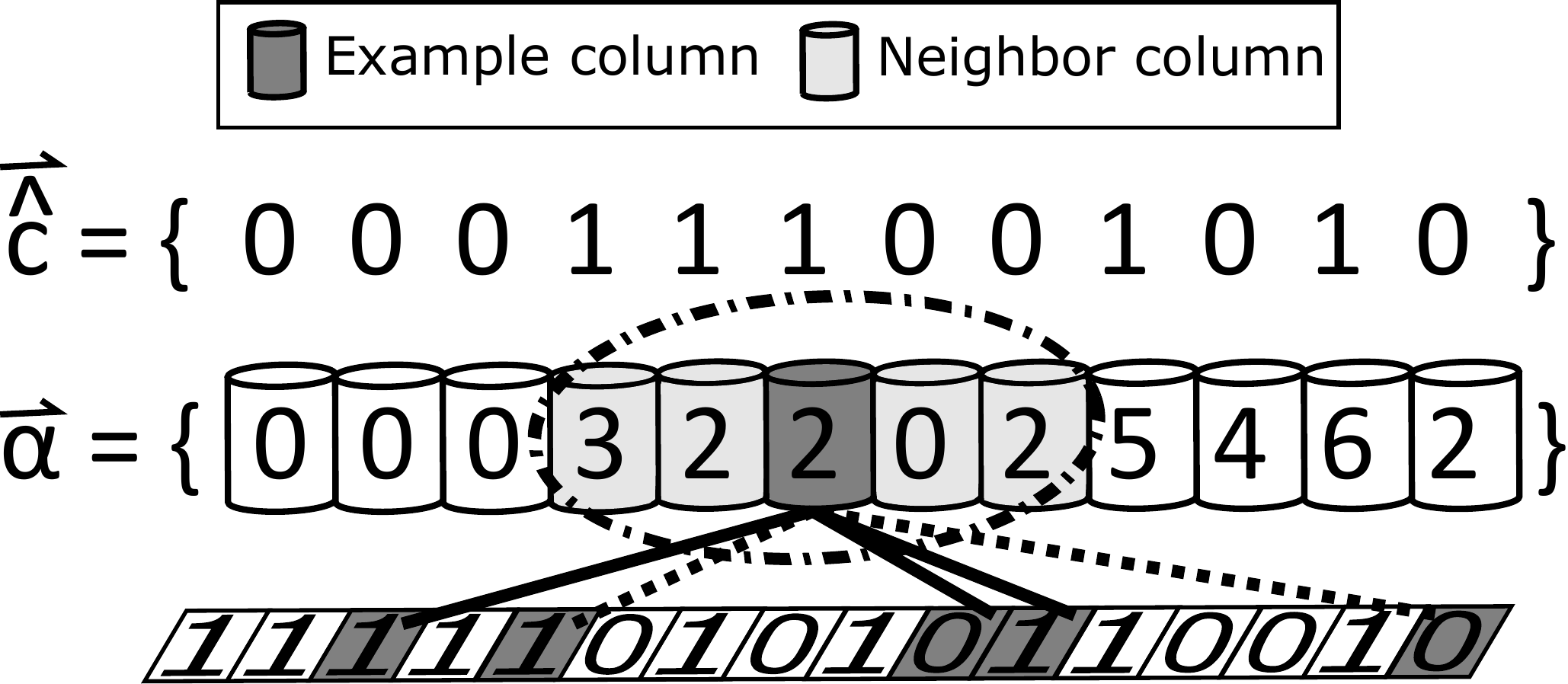}
				\caption{SP phase 2 example where $\rho_c = 2$ and $\sigma_o = 2$. The overlap values were determined from the SP phase 1 example.}
				\label{fig:phase2}
			\end{figure}
			
			\begin{equation}
				\vect{\alpha} \equiv 
				\begin{cases}
					\vect[0.4]{\hat{\alpha}}_i\vect[0.4]{b}_i & \vect[0.4]{\hat{\alpha}}_i \geq \rho_d, \\
					0                 & \operatorname{otherwise}
				\end{cases}
				\ \forall i
				\label{eq:overlap}
			\end{equation}
			
			\begin{equation}
				\vect{\hat{\alpha}}_i \equiv \boldsymbol{X}_i \bullet \boldsymbol{Y}_i
				\label{eq:preoverlap}
			\end{equation}
		
		\subsection{Phase 2: Inhibition}
			Let $\boldsymbol{H} \in \{0, 1\}^{m \times m}$ be defined as the neighborhood mask for all columns, such that $\boldsymbol{H}_i$ is the neighborhood for $\vect{c}_i$. $\vect{c}_j$ is then said to be in $\vect{c}_i$'s neighborhood if and only if $\boldsymbol{H}_{i,j}$ is `1'. Let $\operatorname{kmax}(S, k)$ be defined as the $k$-th largest element of $S$. Let $\operatorname{max}(\vect{v})$ be defined as a function that will return the maximum value in $\vect{v}$. The set of active columns, $\vect[0.3]{\hat{c}} \in \{0, 1\}^{1 \times m}$, may then be obtained by using \eq{eq:inhibition}, where $\vect[0.3]{\hat{c}}$ is an indicator vector representing the activation (or lack of activation) for each column. The result of the indicator function is determined by $\vect{\gamma} \in \mathbb{Z}^{1 \times m}$, which is defined in \eq{eq:activity_level} as the $\rho_c$-th largest overlap (lower bounded by one) in the neighborhood of $\vect{c}_i\ \forall i$.
			
			Comparing these equations with \alg{alg:phase2}, $\vect{\gamma}$ is a slightly altered version of $mo$. Instead of just being the $\rho_c$-th largest overlap for each column, it is additionally lower bounded by one. Referring back to \alg{alg:phase2}, line 3 is a biconditional statement evaluating to true if the overlap is at least $mo$ and greater than zero. By simply enforcing $mo$ to be at least one, the biconditional is reduced to a single condition. That condition is evaluated within the indicator function; therefore, \eq{eq:inhibition} carries out the logic in the if statement in \alg{alg:phase2}. Continuing with the demonstration shown in \fig{fig:phase1}, \fig{fig:phase2} shows an example execution of phase two.
			
			\begin{equation}
				\vect[0.3]{\hat{c}} \equiv \operatorname{I}(\vect{\alpha}_i \geq \vect{\gamma}_i)\ \forall i
				\label{eq:inhibition}
			\end{equation}
			
			\begin{equation}
				\vect{\gamma} \equiv \operatorname{max}(\operatorname{kmax}(\boldsymbol{H}_i \odot \vect{\alpha}, \rho_c), 1)\ \forall i
				\label{eq:activity_level}
			\end{equation}
		
		\subsection{Phase 3: Learning}
			Let $\operatorname{clip}(\boldsymbol{M}, lb, ub)$ be defined as a function that will clip all values in the matrix $\boldsymbol{M}$ outside of the range $[lb, ub]$ to $lb$ if the value is less than $lb$, or to $ub$ if the value is greater than $ub$. $\boldsymbol{\Phi}$ is then recalculated by \eq{eq:permanence_update}, where $\boldsymbol{\delta\Phi}$ is the proximal synapse's permanence update amount given by \eq{eq:permanence_delta}\footnote{Due to $\boldsymbol{X}$ being binary, a bitwise negation is equivalent to the shown logical negation. In a similar manner, the multiplications of $\vect[0.4]{\hat{c}^T}$ with $\boldsymbol{X}$ and $\neg\boldsymbol{X}$ can be replaced by an $\operatorname{AND}$ operation (logical or bitwise).}.
			
			\begin{equation}
				\boldsymbol{\Phi} \equiv \operatorname{clip}\left(\boldsymbol{\Phi} \oplus \boldsymbol{\delta\Phi}, 0, 1\right)
				\label{eq:permanence_update}
			\end{equation}
			
			\begin{equation}
				\boldsymbol{\delta\Phi} \equiv \vect[0.4]{\hat{c}}^T \odot (\phi_+\boldsymbol{X} - (\phi_-\neg\boldsymbol{X}))
				\label{eq:permanence_delta}
			\end{equation}
			
\begin{figure}[!t]
				\centering
				\includegraphics[width=\linewidth]{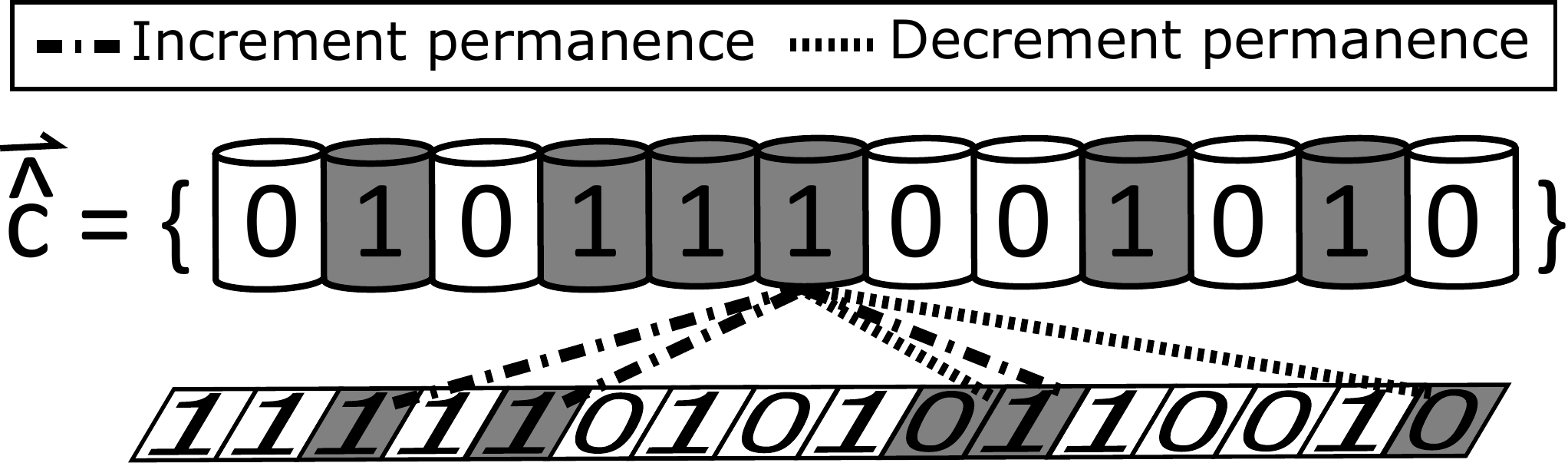}
				\caption{SP phase 3 example, demonstrating the adaptation of the permanences. The gray columns are used denote the active columns, where those activations were determined from the SP phase 2 example.}
				\label{fig:phase3}
			\end{figure}

			The result of these two equations is equivalent to the result of executing the first seven lines in \alg{alg:phase3}. If a column is active, it will be denoted as such in \vect[0.3]{\hat{c}}; therefore, using that vector as a mask, the result of \eq{eq:permanence_delta} will be a zero if the column is inactive, otherwise it will be the update amount. From \alg{alg:phase3}, the update amount should be $\phi_+$ if the synapse was active and $\phi_-$ if the synapse was inactive. A synapse is active only if its source column is active. That activation is determined by the corresponding value in $\boldsymbol{X}$. In this manner, $\boldsymbol{X}$ is also being used as a mask, such that active synapses will result in the update equalling $\phi_+$ and inactive synapses (selected by inverting $\boldsymbol{X}$) will result in the update equalling $\phi_-$. By clipping the element-wise sum of $\boldsymbol{\Phi}$ and $\boldsymbol{\delta\Phi}$, the permanences stay bounded between [0, 1]. As with the previous two phases, the visual demonstration is continued, with \fig{fig:phase3} illustrating the primary functionality of this phase.
			
			Let $\vect[0.3]{\eta^{(a)}} \in \mathbb{R}^{1 \times m}$ be defined as the set of active duty cycles for all columns, such that $\vectmd[0.3]{\boldsymbol{\eta}_i^{\boldsymbol{(a)}}}$ is the active duty cycle for $\vect{c_i}$. Let $\vect[0.3]{\eta^{(min)}} \in \mathbb{R}^{1 \times m}$ be defined by \eq{eq:min_duty_cycle} as the set of minimum active duty cycles for all columns, such that $\vectmd[0.3]{\boldsymbol{\eta}_i^{\boldsymbol{(min)}}}$ is the minimum active duty cycle for $\vect{c}_i$. This equation is clearly the same as line 9 in \alg{alg:phase3}.
			
			\begin{equation}
				\vect[0.3]{\eta^{(min)}} \equiv \kappa_a\operatorname{max}\left(\boldsymbol{H}_i \odot \vect[0.3]{\eta^{(a)}}\right)\ \forall i
				\label{eq:min_duty_cycle}
			\end{equation}
			
			Let $\operatorname{update\_active\_duty\_cycle}(\vect{c})$ be defined as a function that updates the moving average duty cycle for the active duty cycle for each $\vect{c_i} \in \vect{c}$. That function should compute the frequency of each column's activation. After calling $\operatorname{update\_active\_duty\_cycle}(\vect{c})$, the boost for each column is updated by using \eq{eq:boost_update}. In \eq{eq:boost_update}, $\beta\left(\vectmd[0.3]{\boldsymbol{\eta}_i^{\boldsymbol{(a)}}}, \vectmd[0.3]{\boldsymbol{\eta}_i^{\boldsymbol{(min)}}}\right)$ is defined as the boost function, following \eq{eq:boost_calc}\footnote{The conditions within the piecewise function must be evaluated top-down, such that the first condition takes precedence over the second condition which takes precedence over the third condition.}. The functionality of \eq{eq:boost_update} is therefore shown to be equivalent to \alg{alg:boost}.
			
			\begin{equation}
				\vect[0.3]{b} \equiv \beta\left(\vectmd[0.3]{\boldsymbol{\eta}_i^{\boldsymbol{(a)}}}, \vectmd[0.3]{\boldsymbol{\eta}_i^{\boldsymbol{(min)}}}\right) \ \forall i
				\label{eq:boost_update}
			\end{equation}
			
			\begin{equation}
				\resizebox{\hsize}{!}{$%
					\beta\left(\vectmd[0.3]{\boldsymbol{\eta}_i^{\boldsymbol{(a)}}}, \vectmd[0.3]{\boldsymbol{\eta}_i^{\boldsymbol{(min)}}}\right) \equiv %
					\begin{cases}%
						\beta_0 & \vectmd[0.3]{\boldsymbol{\eta}_i^{\boldsymbol{(min)}}} = 0 \\[1ex]%
						1 & \vectmd[0.3]{\boldsymbol{\eta}_i^{\boldsymbol{(a)}}} > \vectmd[0.3]{\boldsymbol{\eta}_i^{\boldsymbol{(min)}}} \\[1ex]%
						\vectmd[0.3]{\boldsymbol{\eta}_i^{\boldsymbol{(a)}}}\frac{1 - \beta_0}{\vphantom{{{{{{a^k}^k}^k}^k}^k}^k}{\vectmd[0.3]{\boldsymbol{\eta}_i^{\boldsymbol{(min)}}}}} + \beta_0 & \operatorname{otherwise}%
					\end{cases}$%
				}
				\label{eq:boost_calc}
			\end{equation}
			
			Let $\vect[0.3]{\eta^{(o)}} \in \mathbb{R}^{1 \times m}$ be defined as the set of overlap duty cycles for all columns, such that $\vectmd[0.3]{\boldsymbol{\eta}_i^{\boldsymbol{(o)}}}$ is the overlap duty cycle for $\vect{c}_i$. Let $\operatorname{update\_overlap\_duty\_cycle}(\vect{c})$ be defined as a function that updates the moving average duty cycle for the overlap duty cycle for each $\vect{c}_i \in \vect{c}$. That function should compute the frequency of each column's overlap being at least equal to $\rho_d$. After applying $\operatorname{update\_overlap\_duty\_cycle}(\vect{c})$, the permanences are then boosted by using \eq{eq:boost_permanence}. This equation is equivalent to lines 13 -- 15 in \alg{alg:phase3}, where the multiplication with the indicator function is used to accomplish the conditional and clipping is done to ensure the permanences stay within bounds.
				
			\begin{equation}
				\boldsymbol{\Phi} \equiv \operatorname{clip}\left(\boldsymbol{\Phi} \oplus \kappa_b\rho_s\operatorname{I}\left(\vectmd[0.3]{\boldsymbol{\eta}_i^{\boldsymbol{(o)}}} < \vectmd[0.3]{\boldsymbol{\eta}_i^{\boldsymbol{(min)}}}\right), 0, 1\right)
				\label{eq:boost_permanence}
			\end{equation}
		
			Let $\operatorname{d}(x, y)$ be defined as the distance function\footnote{The distance function is typically the Euclidean distance.} that computes the distance between $x$ and $y$. To simplify the notation\footnote{In an actual system the positions would be explicitly defined.}, let $\operatorname{pos}(c, r)$ be defined as a function that will return the position of the column indexed at $c$ located $r$ regions away from the current region. For example, $\operatorname{pos}(0, 0)$ returns the position of the first column located in the SP and $\operatorname{pos}(0, -1)$ returns the position of the first column located in the previous region. The distance between $\operatorname{pos}(\vect{c}_i, 0)$ and pos$(\boldsymbol{\Lambda}_{i,k}, -1)$ is then determined by $\operatorname{d}(\operatorname{pos}(\vect{c}_i, 0), \operatorname{pos}(\boldsymbol{\Lambda}_{i,k}, -1))$.
			
			Let $\boldsymbol{D} \in \mathbb{R}^{m \times q}$ be defined as the distance between an SP column and its corresponding connected synapses' source columns, such that $\boldsymbol{D}_{i,k}$ is the distance between $\vect{c}_i$ and $\vect{c}_i$'s proximal synapse's input at index $k$. $\boldsymbol{D}$ is computed following \eq{eq:distances}, where $\boldsymbol{Y}_i$ is used as a mask to ensure that only connected synapses may contribute to the distance calculation. The result of that element-wise multiplication would be the distance between the two columns or zero for connected and unconnected synapses, respectively\footnote{It is assumed that an SP column and an input column do not coincide, i.e. their distance is greater than zero. If this occurs, $\boldsymbol{D}$ will be unstable, as zeros will refer to both valid and invalid distances. This instability is accounted for during the computation of the inhibition radius, such that it will not impact the actual algorithm.}.
			
			\begin{equation}
				\boldsymbol{D} \equiv (\operatorname{d}(\operatorname{pos}(\vect{c}_i, 0), \operatorname{pos}(\boldsymbol{\Lambda}_{i,k}, -1)) \odot \boldsymbol{Y}_i\ \forall k)\ \forall i
				\label{eq:distances}
			\end{equation}
			
			The inhibition radius, $\sigma_0$, is defined by \eq{eq:inhibition_radius}. The division in \eq{eq:inhibition_radius} is the sum of the distances divided by the number of connected synapses\footnote{The summation of the connected synapses is lower-bounded by one to avoid division by zero.}. That division represents the average distance between connected synapses' source and destination columns, and is therefore the average receptive field size. The inhibition radius is then set to the average receptive field size after it has been floored and raised to a minimum value of one, ensuring that the radius is an integer at least equal to one. Comparing \eq{eq:inhibition_radius} to line 16 in \alg{alg:phase3}, the two are equivalent.
			
			\begin{equation}
				\sigma_o \equiv \operatorname{max} \left(1, \left\lfloor\frac{\sum_{i=0}^{m-1}\sum_{k=0}^{q-1}\boldsymbol{D}_{i,k}}{\operatorname{max}(1, \sum_{i=0}^{m-1}\sum_{k=0}^{q-1}\boldsymbol{Y}_{i,k})}\right\rfloor\right)
				\label{eq:inhibition_radius}
			\end{equation}
			
			Once the inhibition radius has been computed, the neighborhood for each column must be updated. This is done using the function $\operatorname{h}(\vect{c}_i)$, which is dependent upon the type of inhibition being used (global or local) as well as the topology of the system\footnote{For global inhibition, every value in $\boldsymbol{H}$ would simply be set to one regardless of the topology. This allows for additional optimizations of \eq{eq:activity_level} and \eq{eq:min_duty_cycle} and eliminates the need for \eq{eq:inhibition_radius} and \eq{eq:neighborhood}. For simplicity only the generalized forms of the equations were shown.}. This function is shown in \eq{eq:neighborhood}, where $\vect[0.3]{\zeta}$ represents all of the columns located at the set of integer Cartesian coordinates bounded by an $n$-dimensional shape. Typically the $n$-dimensional shape is a represented by an $n$-dimensional hypercube.
					
			\begin{equation}
				\operatorname{h}(\vect{c_i}) \equiv 
				\begin{cases}
					\vect{c} & global~inhibition \\
					\vect{\zeta} & local~inhibition
				\end{cases}
				\label{eq:neighborhood}
			\end{equation}
	
	\section{Boosting}
		\begin{figure}[!t]
			\centering
			\includegraphics[width=\linewidth]{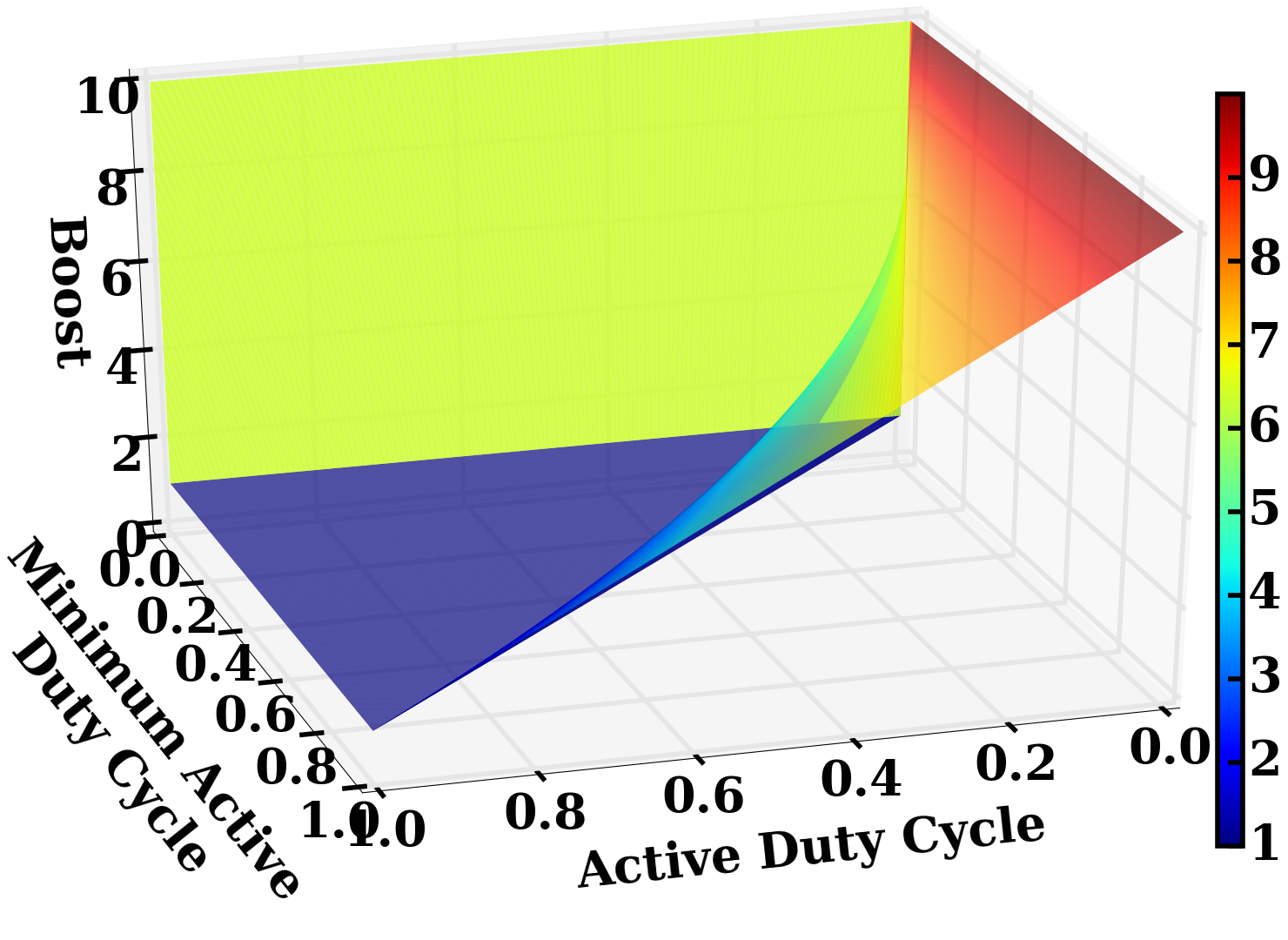}
			\caption{Demonstration of boost as a function of a column's minimum active duty cycle and active duty cycle.}
			\label{fig:boost_value}
		\end{figure}
				
		It is important to understand the dynamics of boosting utilized by the SP. The SP's boosting mechanism is similar to DeSieno's~\cite{desieno} conscience mechanism. In that work, clusters that were too frequently active were penalized, allowing weak clusters to contribute to learning. The SP's primary boosting mechanism takes the reverse approach by rewarding infrequently active columns. Clearly, the boosting frequency and amount will impact the SP's learned representations.
		
		The degree of activation is determined by the boost function, \eq{eq:boost_calc}. From that equation, it is clear that a column's boost is determined by the column's minimum active duty cycle as well as the column's active duty cycle. Those two values are coupled, as a column's minimum active duty cycle is a function of its duty cycle, as shown in \eq{eq:min_duty_cycle}. To study how those two parameters affect a column's boost, \fig{fig:boost_value} was created. From this plot it is found that the non-boundary conditions for a column's boost follows the shape $1 / \vectmd[0.3]{\boldsymbol{\eta}_i^{\boldsymbol{(min)}}}$. It additionally shows the importance of evaluating the piecewise boost function in order. If the second condition is evaluated before the first condition, the boost will be set to its minimum, instead of its maximum value.
		
\begin{figure}[!t]
			\centering
			\includegraphics[width=\linewidth]{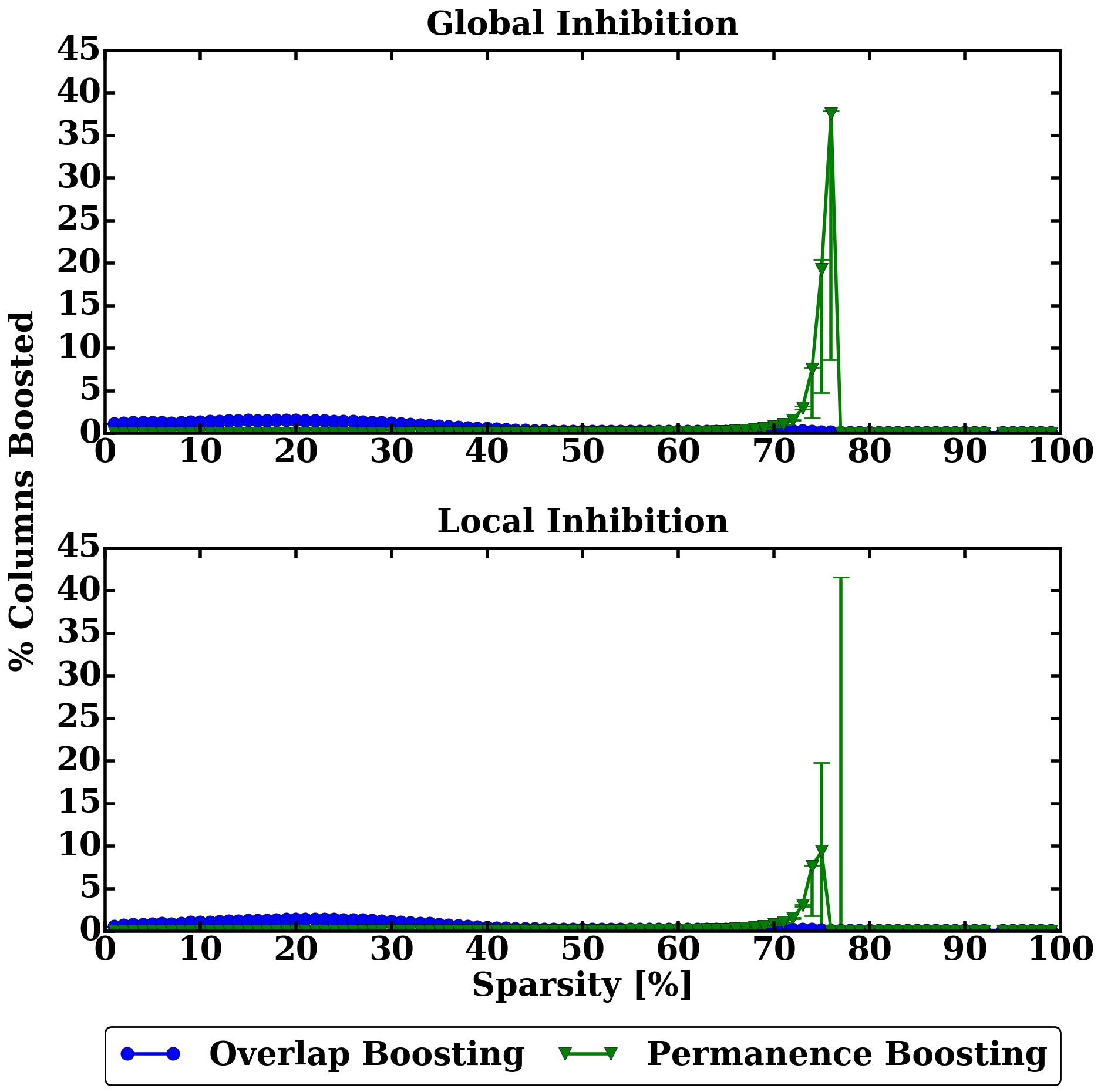}
			\caption{Demonstration of frequency of both boosting mechanisms as a function of the sparseness of the input. The top figure shows the results for global inhibition and the bottom figure shows the results for local inhibition\protect\footnotemark.}
			\label{fig:boost_sparseness}
		\end{figure}

		To study the frequency of boosting, the average number of boosted columns was observed by varying the level of sparseness in the input for both types of inhibition, as shown in \fig{fig:boost_sparseness}. For the overlap boosting mechanism, \eq{eq:boost_update}, very little boosting occurs, with boosting occurring more frequently for denser inputs. This is to be expected, as more bits would be active in the input; thus, causing more competition to occur among the columns.
		
		For the permanence boosting mechanism, \eq{eq:boost_permanence}, boosting primarily occurs when the sparsity is between 70 and 76\%, with almost no boosting occurring outside of that range. That boosting is a result of the SP's parameters. In this experiment, $q = 40$ and $\rho_d = 15$. Based off those parameters, there must be at least 25 active inputs on a column for it have a non-zero overlap; i.e. if the sparsity is 75\%, a column would have to be connected to each active bit in the input to obtain a non-zero overlap. As such, if the sparsity is greater than 75\% it will not be possible for the columns to have a non-zero overlap, resulting in no boosting. For lower amounts of sparsity, boosting was not needed, since adequate input coverage of the active bits is obtained.
		
		Recall that the initialization of the SP is performed randomly. Additionally, the positions of active bits for this dataset are random. That randomness combined with a starved input results in a high degree of volatility. This is observed by the extremely large error bars. Some of the SP's initializations resulted in more favorable circumstances, since the columns were able to connect to the active bits in the input.
		
\begin{figure}[!t]
			\centering
			\includegraphics[width=\linewidth]{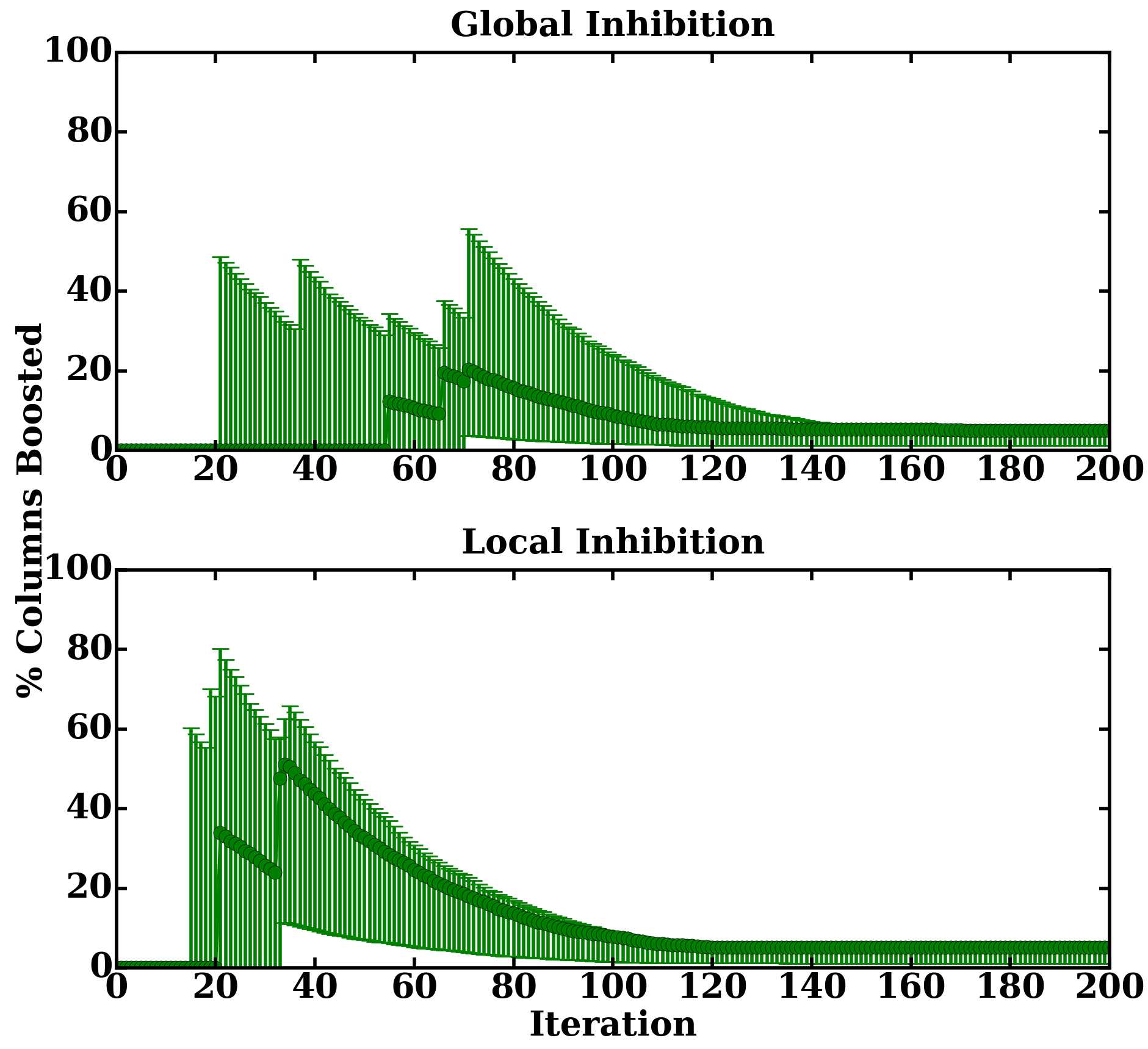}
			\caption{Frequency of boosting for the permanence boosting mechanism for a sparsity of 74\%. The top figure shows the results for global inhibition and the bottom figure shows the results for local inhibition Only the first 200 iterations were shown, for clarity, as the remaining 800 propagated the trend.}
			\label{fig:boost_permanence}
		\end{figure}

		For the SP to adapt to the lack of active bits, it would have to boost its permanence. This would result in a large amount of initial boosting, until the permanences reached a high enough value. Once the permanences reach that value, boosting will only be required occasionally, to ensure the permanences never fully decay. This behavior is observed in \fig{fig:boost_permanence}, where the permanence boosting frequency was plotted for a sparsity of 74\%. The delayed start occurs because the SP has not yet determined which columns will need to be boosted. Once that set is determined, a large amount of boosting occurs. The trend follows a decaying exponential that falls until its minimum level is reached, at which point the overall degree of boosting remains constant. This trend was common among the sparsities that resulted in a noticeable degree of permanence boosting. The right-skewed decaying exponential was also observed in DeSieno's work~\cite{desieno}.
		
		\footnotetext{To obtain this plot, the average number of boosted columns during each epoch is computed. The average, across all of those epochs, is then calculated. That average represents the percentage of columns boosted for a particular level of sparsity. It is important to note that because the SP's synaptic connections are randomly determined, the only dataset specific factors that will affect how the SP performs will be $p$ and the number of bits in the input that are active. This means that it is possible to generalize the SP's behavior for any dataset, provided that the dataset has a relatively constant number of active bits for each pattern. For the purposes of this experiment, the number of active bits was kept fixed, but the positions were varied; thereby creating a generalizable template.
		
		The inputs to the SP consisted of 100 randomly generated bit-streams with a width of 100 bits. Within each bit-stream, bits were randomly flipped to be active. The sparseness is then the percentage of non-active bits. Each simulation consisted of 10 epochs and was performed across 10 trials. The SP's parameters are as follows: $m = 2048$, $p = 100$, $q = 40$, $\rho_d = 15$, $\rho_s = 0.5$, $\phi_{\delta} = 0.05$, $\rho_c = \lfloor 0.02 * m \rfloor$, $\phi_+ = 0.03$, $\phi_- = 0.05$, $\beta_0 = 10$, and $\tau = 100$. Synapses were trimmed if their permanence value ever reached or fell below $10^{-4}$. On the figure, each point represents a partial box plot, i.e. the data point is the median, the upper error bar is the third quartile, and the lower error bar is the first quartile.}
		
		These results show that the need for boosting can be eliminated by simply choosing appropriate values for $q$ and $\rho_d$\footnote{This assumes that there will be enough active bits in the input. If this is not the case, the input may be transformed to have an appropriate level of active bits.}. It is thus concluded that these boosting mechanisms are secondary learning mechanisms, with the primary learning occurring from the permanence update in \eq{eq:permanence_update}. These findings allow resource limited systems (especially hardware designs) to exclude boosting, while still obtaining comparable results; thereby, greatly reducing the complexity of the system.
		
	\section{Feature learning}
		\subsection{Probabilistic Feature Mapping}
			It is convenient to think of a permanence value as a probability. That probability is used to determine if a synapse is connected or unconnected. It also represents the probability that the synapse's input bit is important. It is possible for a given input bit to be represented in multiple contexts, where the context for a specific instance is defined to be the set of inputs connected, via proximal synapses, to a column. Due to the initialization of the network, it is apparent that each context represents a random subspace; therefore, each column is learning the probability of importance for its random subset of attributes in the feature space. This is evident in \eq{eq:permanence_delta}, as permanences contributing to a column's activation are positively reinforced and permanences not contributing to a column's activation are negatively reinforced.
			
			If all contexts for a given input bit are observed, the overall importance of that bit is obtained. Multiple techniques could be conjured for determining how the contexts are combined. The most generous method is simply to observe the maximum. In this manner, if the attribute was important in at least one of the random subspaces, it would be observed. Using those new probabilities the degree of influence of an attribute may be obtained. Let $\vect[0.3]{\hat{\phi}} \in (0, 1)^{1 \times p}$ be defined as the set of learned attribute probabilities. One form of $\vect[0.3]{\hat{\phi}}$ is shown in \eq{eq:permanence_map}\footnote{The function $\operatorname{max}$ was used as an example. Other functions producing a valid probability are also valid.}. In \eq{eq:permanence_map}, the indicator function is used to mask the permanence values for $\boldsymbol{U}_{s,r}$. Multiplying that value by every permanence in $\boldsymbol{\Phi}$ obtains all of the permanences for $\boldsymbol{U}_{s,r}$. This process is used to project the SP's representation of the input back into the input space.
			
			\begin{equation}
				\vect[0.3]{\hat{\phi}} \equiv \operatorname{max}\left(\boldsymbol{\Phi}_{i,k}\operatorname{I}(\boldsymbol{\Lambda}_{i,k}=r)\ \forall i\ \forall k\right)\ \forall r
				\label{eq:permanence_map}
			\end{equation}
		
		\subsection{Dimensionality Reduction}
			The learned attribute probabilities may be used to perform dimensionality reduction. Assuming the form of $\vect[0.3]{\hat{\phi}}$ is that in \eq{eq:permanence_map}, the probability is stated to be important if it is at least equal to $\rho_s$. This holds true, as $\vect[0.3]{\hat{\phi}}$ is representative of the maximum permanence for each input in $\boldsymbol{U}_s$. For a given $\boldsymbol{U}_{s,r}$ to be observed, it must be connected, which may only happen when its permanence is at least equal to $\rho_s$. Given that, the attribute mask, $\vect{z}\in \{0, 1\}^{1 \times p}$, is defined to be $\operatorname{I}\left(\vect[0.3]{\hat{\phi}}\ge \rho_s\right)$. The new set of attributes are those whose corresponding index in the attribute mask are true, i.e. $\boldsymbol{U}_{s,r}$ is a valid attribute if $\vect{z}_r$ is true.
		
		\subsection{Input Reconstruction}
			Using a concept similar to the probabilistic feature mapping technique, it is possible to obtain the SP's learned representation of a specific pattern. To reconstruct the input pattern, the SP's active columns for that pattern must be captured. This is naturally done during inhibition, where $\vect[0.3]{\hat{c}}$ is constructed. $\vect[0.3]{\hat{c}}$, a function of $\boldsymbol{U}_s$, is used to represent a specific pattern in the context of the SP.
			
			Determining which permanences caused the activation is as simple as using $\vect[0.3]{\hat{c}}$ to mask $\boldsymbol{\Phi}$. Once that representation has been obtained, the process follows that of the probabilistic feature mapping technique, where $\operatorname{I}(\boldsymbol{\Lambda}_{i,k}=r)$ is used as a second mask for the permanences. Those steps will produce a valid probability for each input bit; however, it is likely that there will exist probabilities that are not explicitly in \{0, 1\}. To account for that, the same technique used for dimensionality reduction is applied, by simply thresholding the probability at $\rho_s$. This process is shown in \eq{eq:input_reconstruction}\footnote{The function $\operatorname{max}$ was used as an example. If a different function is utilized, it must be ensured that a valid probability is produced. If a sum is used, it could be normalized; however, if caution is not applied, thresholding with respect to $\rho_s$ may be invalid and therefore require a new thresholding technique.}, where $\vect[0.3]{\hat{u}}\in\{0, 1\}^{1 \times p}$ is defined to be the reconstructed input.
			
			\begin{equation}
				\vect{\hat{u}} \equiv \operatorname{I}\left(\left[\operatorname{max}\left(\boldsymbol{\Phi}_{i,k}\vect[0.3]{\hat{c}}_i\operatorname{I}(\boldsymbol{\Lambda}_{i,k}=r)\ \forall i\ \forall k\right)\ge \rho_s\right]\right)\ \forall r
				\label{eq:input_reconstruction}
			\end{equation}
	
	\section{Experimental Results and Discussion}
		To empirically investigate the performance of the SP, a Python implementation of the SP was created, called math HTM (mHTM)\footnote{This implementation has been released under the MIT license and is available at: https://github.com/tehtechguy/mHTM.}. The SP was tested on both spatial data as well as categorical data. The details of those experiments are explained in the ensuing subsections.
		
		\subsection{Spatial Data}
			\begin{table}[!t]
				\caption[SP performance on MNIST using global inhibition]{SP performance on MNIST using global inhibition\protect\footnotemark}
				\label{tbl:mnist_global}
				\centering
				\begin{tabular}{l|l}
					\hline
					\multicolumn{1}{c|}{Method} & \multicolumn{1}{c}{Error} \\
					\hline
					column & 7.70\% \\
					probabilistic & 8.98\% \\
					reduction & 9.03\% \\
					\hline
				\end{tabular}
			\end{table}
			\footnotetext{The following parameters were used to obtain these results: $m = 936$, $q = 353$, $\rho_d = 14$, $\phi_{\delta} = 0.0105$, $\rho_c = 182$, $\phi_+ = 0.0355$, $\phi_- = 0.0024$, $\beta_0 = 18$, and $\tau = 164$.}
			
			The SP is a spatial algorithm, as such, it should perform well with inherently spatial data. To investigate this, the SP was tested with a well-known computer vision task. The SP requires a binary input; as such, it was desired to work with images that were originally black and white or could be readily made black and white without losing too much information. Another benefit of using this type of image is that the encoder\footnote{An encoder for HTM is any system that takes an arbitrary input and maps it to a new domain (whether by lossy or lossless means) where all values are mapped to the set \{0, 1\}.} may be de-emphasized, allowing for the primary focus to be on the SP. With those constraints, the modified~National~Institute~of~Standards~and~Technology's~(MNIST's) database of handwritten digits~\cite{mnist} was chosen as the dataset.
			
			The MNIST images are simple $28\times28$ grayscale images, with the bulk of the pixels being black or white. To convert them to black and white images, each pixel was set to `1' if the value was greater than or equal to $\nicefrac{255}{2}$ and `0' otherwise. Each image was additionally transformed to be one-dimensional by horizontally stacking the rows. The SP has a large number of parameters, making it difficult to optimize the parameter selection. To help with this, $1,000$ independent trials were created, all having a unique set of parameters. The parameters were randomly selected within reasonable limits\footnote{The following parameters were kept constant: $\rho_s=0.5$, $30$ training epochs, and synapses were trimmed if their permanence value ever reached or fell below $10^{-4}$.}. Additionally, parameters were selected such that $\mathbb{E}[\lambda'] = 0$. To reduce the computational load, the size of the MNIST dataset was reduced to $800$ training samples and $200$ testing samples. The samples were chosen from their respective initial sets using a stratified shuffle split with a total of five splits. To ensure repeatability and to allow proper comparisons, care was taken to ensure that both within and across experiments the same random initializations were occurring. To perform the classification, a linear support vector machine (SVM) was utilized. The input to the SVM was the corresponding output of the SP.
			
			Three comparisons were explored for both global and local inhibition: using the set of active columns as the features (denoted as ``column''), using $\vect[0.3]{\hat{\phi}}$ as the features (denoted as ``probabilistic''), and using the dimensionality reduced version of the input as the features (denoted as ``reduction''). For each experiment the average error of the five splits was computed for each method. The top $10$ performers for each method were then retrained on the full MNIST dataset. From those results, the set of parameters with the lowest error across the experiments and folds was selected as the optimal set of parameters.
			
			\begin{table}[!t]
				\caption[SP performance on MNIST using local inhibition]{SP performance on MNIST using local inhibition\protect\footnotemark}
				\label{tbl:mnist_local}
				\centering
				\begin{tabular}{l|l}
					\hline
					\multicolumn{1}{c|}{Method} & \multicolumn{1}{c}{Error} \\
					\hline
					column & 7.85\% \\
					probabilistic & 9.07\% \\
					reduction & 9.07\% \\
					\hline
				\end{tabular}
			\end{table}\footnotetext{The following parameters were used to obtain these results: $m = 786$, $q = 267$, $\rho_d = 10$, $\phi_{\delta} = 0.0425$, $\rho_c = 57$, $\phi_+ = 0.0593$, $\phi_- = 0.0038$, $\beta_0 = 19$, and $\tau = 755$.}
			
			The results are shown in \tbl{tbl:mnist_global} and \tbl{tbl:mnist_local} for global and local inhibition, respectively. For reference, the same SVM without the SP resulted in an error of 7.95\%. The number of dimensions was reduced by 38.27\% and 35.71\% for global and local inhibition, respectively. Both the probabilistic and reduction methods only performed marginally worse than the base SVM classifier. Considering that these two techniques are used to modify the raw input, it is likely that the learned features were the face of the numbers (referring to inputs equaling `1'). In that case, those methods would almost act as pass through filters, as the SVM is already capable of determining which features are more / less significant. That being said, being able to reduce the number of features by over two thirds, for the local inhibition case, while still performing relatively close to the case where all features are used is quite desirable.
			
			Using the active columns as the learned feature is the default behavior, and it is those activations that would become the feedforward input to the next level (assuming an HTM with multiple SPs and / or TMs). Both global and local inhibition outperformed the SVM, but only by a slight amount. Considering that only one SP region was utilized, that the SP's primary goal is to map the input into a new domain to be understood by the TM, and that the SP did not hurt the SVM's ability to classify, the SP's overall performance is acceptable. It is also possible that given a two-dimensional topology and restricting the initialization of synapses to a localized radius may improve the accuracy of the network. Comparing global to local inhibition, comparable results are obtained. This is likely due to the globalized formation of synaptic connections upon initialization, since that results in a loss of the initial network topology.
			
\begin{figure}[!t]
				\centering
				\includegraphics[width=\linewidth]{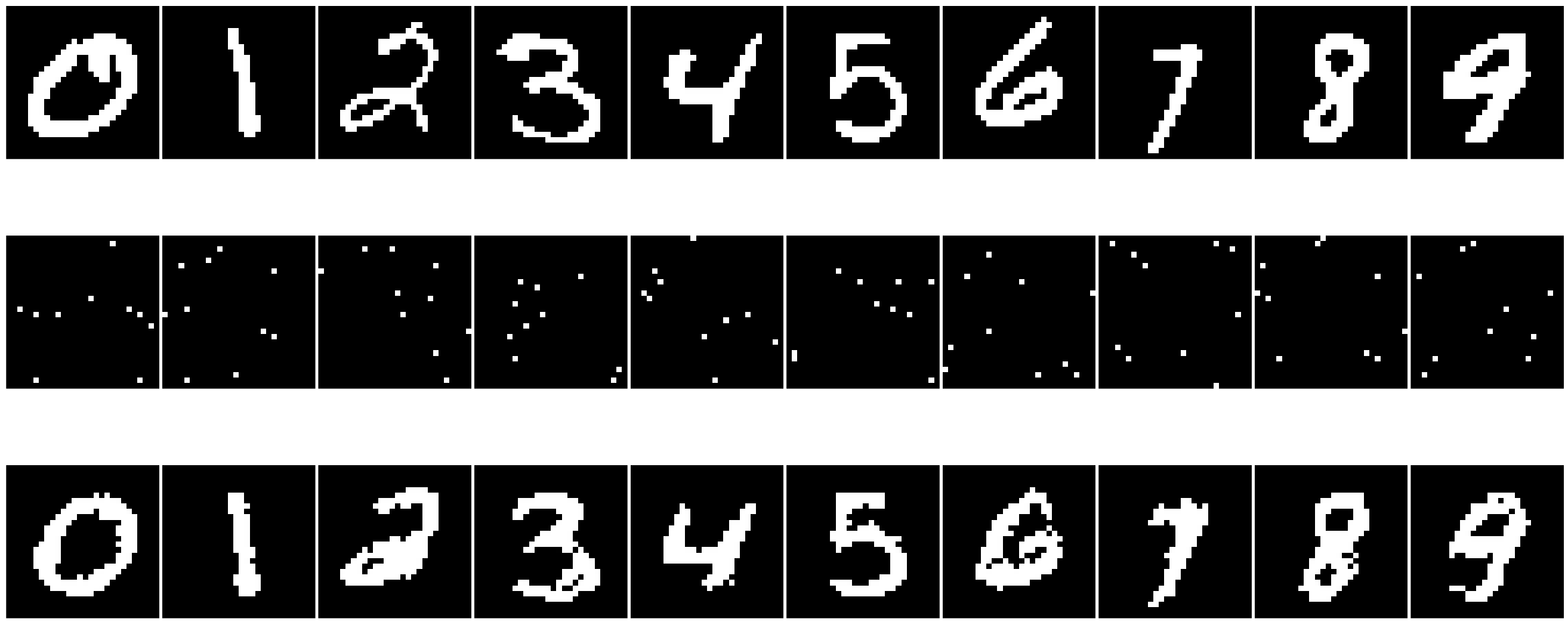}
				\caption[Reconstruction of the input from the context of the SP. Shown are the original input images (top), the sparse distributed representations (middle), and the reconstructed version (bottom).]{Reconstruction of the input from the context of the SP. Shown are the original input images (top), the SDRs (middle), and the reconstructed version (bottom).\protect\footnotemark}
				\label{fig:input_reconstruction}
			\end{figure}

			To explore the input reconstruction technique, a random instance of each class from MNIST was selected. The input was then reconstructed as shown in \fig{fig:input_reconstruction}. The top row shows the original representation of the inputs. The middle row shows the SDR of the inputs. The bottom row shows the reconstructed versions. The representations are by no means perfect, but it is evident that the SP is indeed learning an accurate representation of the input.
			
			\footnotetext{The following parameters were used to obtain these results: $m = 784$, $q = 392$, $\rho_d = 10$, $\phi_{\delta} = 0.01$, $\rho_c = 10$, $\phi_+ = 0.001$, $\phi_- = 0.002$, ten training epochs, global inhibition, and boosting was disabled. The number of columns was set to be equal to the number of inputs to allow for a 1:1 reconstruction of the SDRs.}
		
		\subsection{Categorical Data}
			One of the main purposes of the SP is to create a spatial representation of its input through the process of mapping its input to SDRs. To explore this, the SP was tested on Bohanec et al.'s car evaluation dataset~\cite{bohanec1988knowledge},~\cite{Lichman:2013}. This dataset consists of four classes and six attributes. Each attribute has a finite number of states with no missing values. To encode the attributes, a multivariate encoder comprised of categorical encoders was used\footnote{A multivariate encoder is one which combines one or more other encoders. The multivariate encoder's output concatenates the output of each of the other encoders to form one SDR. A categorical encoder is one which losslessly converts an item to a unique SDR. To perform this conversion, the number of categories must be finite.
			
			For this experiment, each category encoder was set to produce an SDR with a total of 50 bits. The number of categories, for each encoder, was dynamically determined. This value was set to the number of unique instances for each attribute/class. No active bits were allowed to overlap across encodings. The number of active bits, for each encoding, was scaled to be the largest possible value. That scaling would result in utilizing as many of the 50 bits as possible, across all encodings. All encodings have the same number of active bits. In the event that the product of the number of categories and the number of active bits is less than the number of bits, the output was right padded with zeros.}. The class labels were also encoded, by using a single category encoder\footnote{This encoding followed the same process as the category encoders used for the attributes.}.
			
			The selection of the SP's parameters was determined through manual experimentation\footnote{The following parameters were used: $m = 4096$, $q = 25$, $\rho_d = 0$, $\phi_{\delta} = 0.5$, $\rho_c = 819$, $\phi_+ = 0.001$, and $\phi_- = 0.001$. Boosting was disabled and global inhibition was used. Only a single training epoch was utilized. It was found that additional epochs were not required and could result in overfitting. $\rho_c$ was intentionally set to be about 20\% of $m$. This deviates from the standard value of $\sim2\%$. A value of 2\% resulted in lower accuracies (across many different parameter settings) than 20\%. This result is most likely a result of the chosen classifier. For use with the TM or a different classifier, additional experimentation will be required.}. Cross validation was used, by partitioning the data using a stratified shuffle split with eight splits. To perform the classification an SVM was utilized, where the output of the SP was the corresponding input the SVM. The SP's performance was also compared to just using the linear SVM and using a random forest classifier\footnote{These classifiers were utilized from scikit-learn~\cite{scikit}.}. For those classifiers, a simple preprocessing step was performed to map the text-based values to integers.
			
			\begin{table}[!t]
				\caption{Comparison of classifiers on car evaluation dataset}
				\label{tbl:car_eval}
				\centering
				\begin{tabular}{l|l}
					\hline
					\multicolumn{1}{c|}{Classifier} & \multicolumn{1}{c}{Error} \\
					\hline
					Linear SVM & 26.01\% \\
					Random Forest & 8.96\% \\
					SP + Linear SVM & 1.73\% \\
					\hline
				\end{tabular}
			\end{table}
			
			The results are shown in \tbl{tbl:car_eval}\footnote{The shown error is the median across all splits of the data.}. The SVM performed poorly, having an error of 26.01\%. Not surprisingly, the random forest classifier performed much better, obtaining an error of 8.96\%. The SP was able to far outperform either classifier, obtaining an error of only 1.73\%. From literature, the best known error on this dataset was 0.37\%, which was obtained from a boosted multilayer perceptron~(MLP)~\cite{oza2005online}. Comparatively, the SP with the SVM is a much less complicated system.
			
			This result shows that the SP was able to map the input data into a suitable format for the SVM, thereby drastically improving the SVM's classification. Based off this, it is determined that the SP produced a suitable encoding.
			
		\subsection{Extended Discussion}	
			Comparing the SP's performance on spatial data to that of categorical data provides some interesting insights. It was observed that on spatial data the SP effectively acted as a pass through filter. This behavior occurs because the data is inherently spatial. The SP maps the spatial data to a new spatial representation. This mapping allows classifiers, such as an SVM, to be able to classify the data with equal effectiveness.
			
			Preprocessing the categorical data with the SP provided the SVM with a new spatial representation. That spatial representation was understood by the SVM as readily as if it were inherently spatial. This implies that the SP may be used to create a spatial representation from non-spatial data. This would thereby provide other algorithms, such as the TM and traditional spatial classifiers, a means to interpret non-spatial data.
	
	\section{Exploring the Primary Learning Mechanism}		
		To complete the mathematical formulation it is necessary to define a function governing the primary learning process. Within the SP, there are many learned components: the set of active columns, the neighborhood (through the inhibition radius), and both of the boosting mechanisms. All of those components are a function of the permanence, which serves as the probability of an input bit being active in various contexts.
		
		As previously discussed, the permanence is updated by \eq{eq:permanence_delta}. That update equation may be split into two distinct components. The first component is the set of active columns, which is used to determine the set of permanences to update. The second component is the remaining portion of that equation, and is used to determine the permanence update amount.
		
		\subsection{Plausible Origin for the Permanence Update Amount}
			In the permanence update equation, \eq{eq:permanence_delta}, it is noted that the second component is an unlearned function of a random variable coming from a prior distribution. That random variable is nothing more than $\boldsymbol{X}$. It is required that $\boldsymbol{X}_{i,k} \sim \operatorname{Ber}(\mathbb{P}(\boldsymbol{X}_{i,k}))$, where $\operatorname{Ber}$ is used to denote the Bernoulli distribution. If it is assumed that each $\boldsymbol{X}_{i,k} \in \boldsymbol{X}$ are independent and  identically distributed (i.i.d.), then $\boldsymbol{X} \distas{i.i.d.} \operatorname{Ber}(\theta)$, where $\theta$ is defined to be the probability of an input being active. Using the PMF of the Bernoulli distribution, the likelihood of $\theta$ given $\boldsymbol{X}$ is obtained in \eq{eq:joint_likelihood}, where $t\equiv mq$ and $\boldsymbol{\overline{X}} \equiv \frac{1}{t}\sum_{i=0}^{m-1}\sum_{k=0}^{q-1}\boldsymbol{X}_{i,k}$, the overall mean of $\boldsymbol{X}$. The corresponding log-likelihood of $\theta$ given $\boldsymbol{X}$ is given in \eq{eq:joint_log_likelihood}.
			
			\begin{equation}
				\begin{split}
					\mathcal{L}(\theta; \boldsymbol{X}) &= \prod_{i=0}^{m}\prod_{k=0}^{q}\theta^{\boldsymbol{X}_{i,k}}(1-\theta)^{1-\boldsymbol{X}_{i,k}}\\
					&=\theta^{t\boldsymbol{\overline{X}}}(1-\theta)^{t-t\boldsymbol{\overline{X}}}
				\end{split}
				\label{eq:joint_likelihood}
			\end{equation}
			
			\begin{equation}
				\ell(\theta; \boldsymbol{X}) = t\boldsymbol{\overline{X}}\text{log}(\theta) + (t-t\boldsymbol{\overline{X}})\text{log}(1-\theta)
				\label{eq:joint_log_likelihood}
			\end{equation}
			
			Taking the gradient of the joint log-likelihood of \eq{eq:joint_log_likelihood} with respect to $\theta$, results in \eq{eq:max_joint_log_likelihood}. Ascending that gradient results in obtaining the maximum-likelihood estimator (MLE) of $\theta$, $\hat{\theta}_{MLE}$. It can be shown that $\hat{\theta}_{MLE} = \boldsymbol{\overline{X}}$. In this context, $\hat{\theta}_{MLE}$ is used as an estimator for the maximum probability of an input being active.
			
			\begin{equation}
				\label{eq:max_joint_log_likelihood}
				\boldsymbol{\nabla}\ell(\theta; \boldsymbol{X}) = \frac{t}{\theta}\boldsymbol{\overline{X}} - \frac{t}{1-\theta}(1-\boldsymbol{\overline{X}})
			\end{equation}
			
			Taking the partial derivative of the log-likelihood for a single $\boldsymbol{X}_{i,k}$ results in \eq{eq:max_log_likelihood}. Substituting out $\theta$ for its estimator, $\boldsymbol{\overline{X}}$, and multiplying by $\kappa$, results in \eq{eq:p_update_a}. $\kappa$ is defined to be a scaling parameter and must be defined such that $\frac{\kappa}{\boldsymbol{\overline{X}}} \in [0, 1]$ and $\frac{\kappa}{1-\boldsymbol{\overline{X}}} \in [0, 1]$. Revisiting the permanence update equation, \eq{eq:permanence_delta}, the permanence update amount is equivalently rewritten as $\phi_+\boldsymbol{X} - \phi_-(\boldsymbol{J}-\boldsymbol{X})$, where $\boldsymbol{J} \in \{1\}^{m \times q}$. For a single $\boldsymbol{X}_{i,k}$ it is clear that the permanence update amount reduces to $\phi_+\boldsymbol{X}_{i,k} - \phi_-(1-\boldsymbol{X}_{i,k})$. If $\phi_+\equiv\frac{\kappa}{\boldsymbol{\overline{X}}}$ and $\phi_-\equiv\frac{\kappa}{1-\boldsymbol{\overline{X}}}$, then \eq{eq:p_update_a} becomes \eq{eq:p_update_b}. Given this, $\boldsymbol{\delta\Psi}$ is presented as a plausible origin for the permanence update amount. Using the new representations of $\phi_+$ and $\phi_-$, a relationship between the two is obtained, requiring that only one parameter, $\kappa$, be defined. Additionally, it is possible that there exists a $\kappa$ such that $\phi_+$ and $\phi_-$ may be optimally defined for the desired set of parameters.
			
			\begin{equation}
				\frac{\partial}{\partial\theta}\ell(\theta; \boldsymbol{X}_{i,k}) = \frac{1}{\theta}\boldsymbol{X}_{i,k} - \frac{1}{1-\theta}(1-\boldsymbol{X}_{i,k})
				\label{eq:max_log_likelihood}
			\end{equation}
			
			\begin{subequations}
				\label{eq:p_update}
				\begin{align}
					\label{eq:p_update_a}
					\boldsymbol{\delta\Psi}_{i,k} &\equiv \frac{\kappa}{\boldsymbol{\overline{X}}}\boldsymbol{X}_{i,k} - \frac{\kappa}{1-\boldsymbol{\overline{X}}}(1-\boldsymbol{X}_{i,k})\\
					\label{eq:p_update_b}
					&\equiv \phi_+\boldsymbol{X}_{i,k} - \phi_-(1-\boldsymbol{X}_{i,k})
				\end{align}
			\end{subequations}
		
		\subsection{Discussing the Permanence Selection}
			The set of active columns is the learned component in \eq{eq:permanence_delta}, obtained through a process similar to competitive learning~\cite{competitive_learning}. In a competitive learning network, each neuron in the competitive learning layer is fully connected to each input neuron. The neurons in the competitive layer then compete, with one neuron winning the competition. The neuron that wins sets its output to `1' while all other neurons set their output to `0'. At a global scale, this resembles the SP with two key differences. The SP permits multiple columns to be active at a time and each column is connected to a different subset of the input.
			
			Posit that each column is equivalent to a competitive learning network. This would create a network with one neuron in the competitive layer and $q$ neurons in the input layer. The neuron in the competitive layer may only have the state of `1' or `0'; therefore, only one neuron would be active at a time. Given this context, each column is shown to follow the competitive learning rule.
			
			Taking into context the full SP, with each column as a competitive learning network, the SP could be defined to be a bag of competitive learning networks, i.e. an ensemble with a type of competitive learning network as its base learner. Recalling that $\boldsymbol{X} \subseteq \boldsymbol{U}_s$, each $\boldsymbol{X}_i$ is an input for $\vect{c}_i$. Additionally each $\boldsymbol{X}_i$ is obtained by randomly sampling $\boldsymbol{U}_s$ without replacement. Comparing this ensemble to attribute bagging~\cite{attribute_bagging}, the primary difference is that sampling is done without replacement instead of with replacement.
			
			In attribute bagging, a scheme, such as voting, must be used to determine what the result of the ensemble should be. For the SP, a form of voting is performed through the construction of $\vect{\alpha}$. Each base learner (column) computes its degree of influence. The max degree of influence is $q$. Since that value is a constant, each $\vect{\alpha}_i$ may be represented as a probability by simply dividing $\vect{\alpha}_i$ by $q$. In this context, each column is trying to maximize its probability of being selected. During the inhibition phase, a column is chosen to be active if its probability is at least equal to the $\rho_c$-th largest probability in its neighborhood. This process may then be viewed as a form of voting, as all columns within a neighborhood cast their overlap value as their vote. If the column being evaluated has enough votes, it will be placed in the active state.
	
	\section{Conclusion \& Future Work}
		In this work, a mathematical framework for HTM's SP was presented. Using the framework, it was demonstrated how the SP can be used for feature learning. The primary learning mechanism of the SP was explored. It was shown that the mechanism consists of two distinct components, permanence selection and the degree of permanence update. A plausible estimator was provided for determining the degree of permanence update, and insight was given on the behavior of the permanence selection.
		
		The findings in this work provide a basis for intelligently initializing the SP. Due to the mathematical framework, the provided equations could be used to optimize hardware designs. Such optimizations may include removing the boosting mechanism, limiting support to global inhibition, exploiting the matrix operations to improve performance, reducing power through the reduction of multiplexers, etc\ldots. In the future, it is planned to explore optimized hardware designs. Additionally, it is planned to expand this work to provide the same level of understanding for the TM.
	
	\section*{Acknowledgment}
		The authors would like to thank K. Gomez of Seagate Technology, the staff at RIT's research computing, and the members of the NanoComputing Research Lab, in particular A. Hartung and C. Merkel, for their support and critical feedback.
	
	\bibliographystyle{IEEEtran}
		\bibliography{IEEEabrv,sp_math}
\end{document}